# IM-Chat: A Multi-agent LLM Framework Integrating Tool-Calling and Diffusion Modeling for Knowledge Transfer in Injection Molding Industry

Junhyeog Lee [1*], Joon-Young Kim [1,2*], Heekyu Kim [1*], Inhyo Lee [1], Seunghwa Ryu [1‡]

**Affiliations**

[1]Department of Mechanical Engineering, Korea Advanced Institute of Science and Technology (KAIST), 291 Daehak-ro, Yuseong-gu, Daejeon 34141, Republic of Korea

[2]Industrial Intelligence Research Group, AI/DX Center, Institute for Advanced Engineering (IAE), Yongin, Republic of Korea

[*]Corresponding authors' e-mail: ryush@kaist.ac.kr

**Abstract**

The injection molding industry faces critical challenges in preserving and transferring field knowledge, particularly as experienced workers retire and multilingual barriers hinder effective communication. This study introduces IM-Chat, a multi-agent framework based on large language models (LLMs), designed to facilitate knowledge transfer in injection molding. IM-Chat integrates both limited documented knowledge (e.g., troubleshooting tables, manuals) and extensive field data modeled through a data-driven process condition generator that infers optimal manufacturing settings from environmental inputs such as temperature and humidity, enabling robust and context-aware task resolution. By adopting a retrieval-augmented generation (RAG) strategy and tool-calling agents within a modular architecture, IM-Chat ensures adaptability without the need for fine-tuning. Performance was assessed across 100 single-tool and 60 hybrid tasks for GPT-4o, GPT-4o-mini, and GPT-3.5-turbo by domain experts using a 10-point rubric focused on relevance and correctness, and was further supplemented by automated evaluation using GPT-4o guided by a domain-adapted instruction prompt. The evaluation results indicate that more capable models tend to achieve higher accuracy, particularly in complex, tool-integrated scenarios. In addition, compared with the fine-tuned single-agent LLM, IM-Chat demonstrated superior accuracy, particularly in quantitative reasoning, and greater scalability in handling multiple information sources. Overall, these findings demonstrate the viability of multi-agent LLM systems for industrial knowledge workflows and establish IM-Chat as a scalable and generalizable approach to AI-assisted decision support in manufacturing.

# 1 INTRODUCTION

## 1.1 Research Background

Injection molding, a core method for mass production of plastic components, underpins a significant portion of the global manufacturing industry [1–4]. Characterized by high-mix, low-volume production, injection molding demands advanced on-site expertise to manage intricate process settings, ensure production efficiency, and maintain rigorous quality control across a diverse product spectrum [5].

In current practice, on-site operators manually inspect products and iteratively adjust process parameters to maintain quality [6]. This process heavily relies on individual experience and expertise, as external factors such as changes in the external environment, characteristics of injection molding machines, and variations in the physical properties of resins directly impact the final product quality [7,8]. Consequently, operators must acquire extensive domain knowledge to rapidly and accurately respond to such multifaceted influences.

Traditionally, knowledge transfer in the injection molding industry has relied on structured rule sets for addressing specific defects, equipment troubleshooting manuals, and apprenticeship models in which experienced operators directly impart their expertise to less experienced or newly hired workers [9–11]. However, these legacy approaches face growing limitations.

A shrinking and aging workforce increasingly challenges the industry. As skilled veteran operators retire and fewer newcomers enter the field, the implicit knowledge accumulated over decades is at risk of being lost [12]. Furthermore, as a low-margin industry, many injection molding companies have reduced staffing or replaced expert operators with less experienced personnel to lower labor costs [13]. These changes highlight the limitations of traditional apprenticeship-based knowledge transfer and underscore the urgent need for new approaches that can systematize, preserve, and disseminate expert-level process knowledge [14]. Additionally, globalization has further exacerbated the challenge, as the increase in a multilingual workforce creates communication barriers in knowledge sharing [15]. In the absence of mechanisms to translate and deliver operational insights across languages and skill levels, production efficiency is compromised. As such, there is a critical need for knowledge transfer systems that not only preserve expert knowledge but also ensure seamless, real-time communication among diverse operators. To achieve this, the system needs to go beyond simple documentation and adopt innovative approaches that can capture, reason with, and deliver operational expertise in dynamic and interpretable ways. Such advancements are essential for sustaining operational continuity and enhancing competitiveness in an increasingly complex global market.

Recent advancements in large language models (LLMs) present a promising solution to this challenge. LLMs have significantly improved the accuracy of natural language reasoning tasks, enabling them to excel in general applications such as inference, mathematical problem-solving, logical reasoning, and everyday conversations [16–21]. Techniques like chain of thought prompting and in-context learning have further enhanced their reasoning capabilities by guiding models to systematically decompose complex problems and recognize patterns from contextual examples without modifying model parameters [22,23]. These advancements indicate that LLMs have the potential to function as systems that seamlessly communicate with workers and facilitate knowledge transfer through reasoning.

Nevertheless, while LLMs have shown strong general performance, domain-specific applications still face limitations, particularly the problem of hallucination, where models generate plausible but inaccurate content. To mitigate these issues, researchers have developed techniques like fine-tuning [24–26], which allows users to adapt pre-trained models to specific tasks with a labeled dataset, and retrieval-augmented generation (RAG), which integrates external knowledge sources into model prompts to improve factual accuracy [27–30]. Additionally, agent-based approaches have emerged, incorporating APIs and algorithms that enable models to dynamically leverage external tools for more precise task execution [31–33]. Expanding on agent-based approaches, multi-agent systems have gained traction as a promising framework, where multiple specialized agents collaborate within structured workflows to tackle complex problems. These developments reflect a broader shift towards creating more flexible, context-aware AI systems that can reliably handle both general and specialized tasks, paving the way for their application as effective knowledge transfer systems [34,35].

While prior multi-agent LLM frameworks such as HuggingGPT [33] and AutoGen [34] have enabled a general-purpose orchestration, the manufacturing domain poses additional challenges that require the integration of domain-specific knowledge sources such as troubleshooting tables, numerical decision like process parameters, and operator manuals. Successful deployment in industrial environments therefore demands careful curation and structuring of these specialized resources to ensure that the system can provide contextually accurate and practically actionable guidance. To the best of our knowledge, no prior study has applied an LLM agent-based approach for knowledge transfer in the field of injection molding, nor has any study integrated a generative deep learning model (such as a diffusion model) trained on field data as a callable tool within a multi-agent LLM workflow to support quantitative decision-making in manufacturing.

To bridge this gap, this study introduces **IM-Chat** (**I**njection **M**olding **Chat**ting interface), a multi-agent framework that leverages knowledge- and tool-augmented LLMs to address the growing need for effective knowledge transfer in the injection molding industry. To the best of our knowledge, this study is the first work to propose an LLM agent-based approach tailored specifically for this domain. The benefits of integrating LLM agents into knowledge transfer systems can be summarized as follows:

(i) Resilient knowledge preservation: IM-Chat overcomes the limitations of time-consuming and fragile apprenticeship systems, offering a scalable solution in the face of workforce aging and language diversity.

(ii) Contextual adaptability via RAG: By integrating RAG, IM-Chat synthesizes insights from legacy documentation, expert knowledge, and online resources, enabling context-aware and dynamically adaptive knowledge transfer.

(iii) Data-driven precision: IM-Chat incorporates additional AI modules, such as surrogate and generative models, via tool-calling, allowing it to provide accurate, quantitative guidance for production decision-making under varying environmental conditions.

The overall schematic representation of this framework is shown in **Figure 1**.

The subsequent sections of this paper outline the design and validation of IM-Chat, focusing on two key knowledge transfer types: (i) domains with limited field knowledge, where experiential or textual information serves as the primary source, and (ii) domains with extensive field knowledge, where machine learning and generative models facilitate data-driven decision-making. The **Methods** section details the curation of knowledge sources and datasets, the development of LLM workflows, and the integration of generative and surrogate models to enhance adaptability. The **Results** section evaluates IM-Chat's performance, assessing its effectiveness across different knowledge transfer scenarios and its compatibility with various LLMs. Furthermore, it compares IM-Chat with the fine-tuned single-agent baseline in terms of accuracy, latency, and pricing. Finally, the **Discussion** and **Conclusion** sections summarize the study's key contributions, address existing challenges, and explore future directions for leveraging LLM-based workflows as advanced knowledge transfer systems in the injection molding industry.

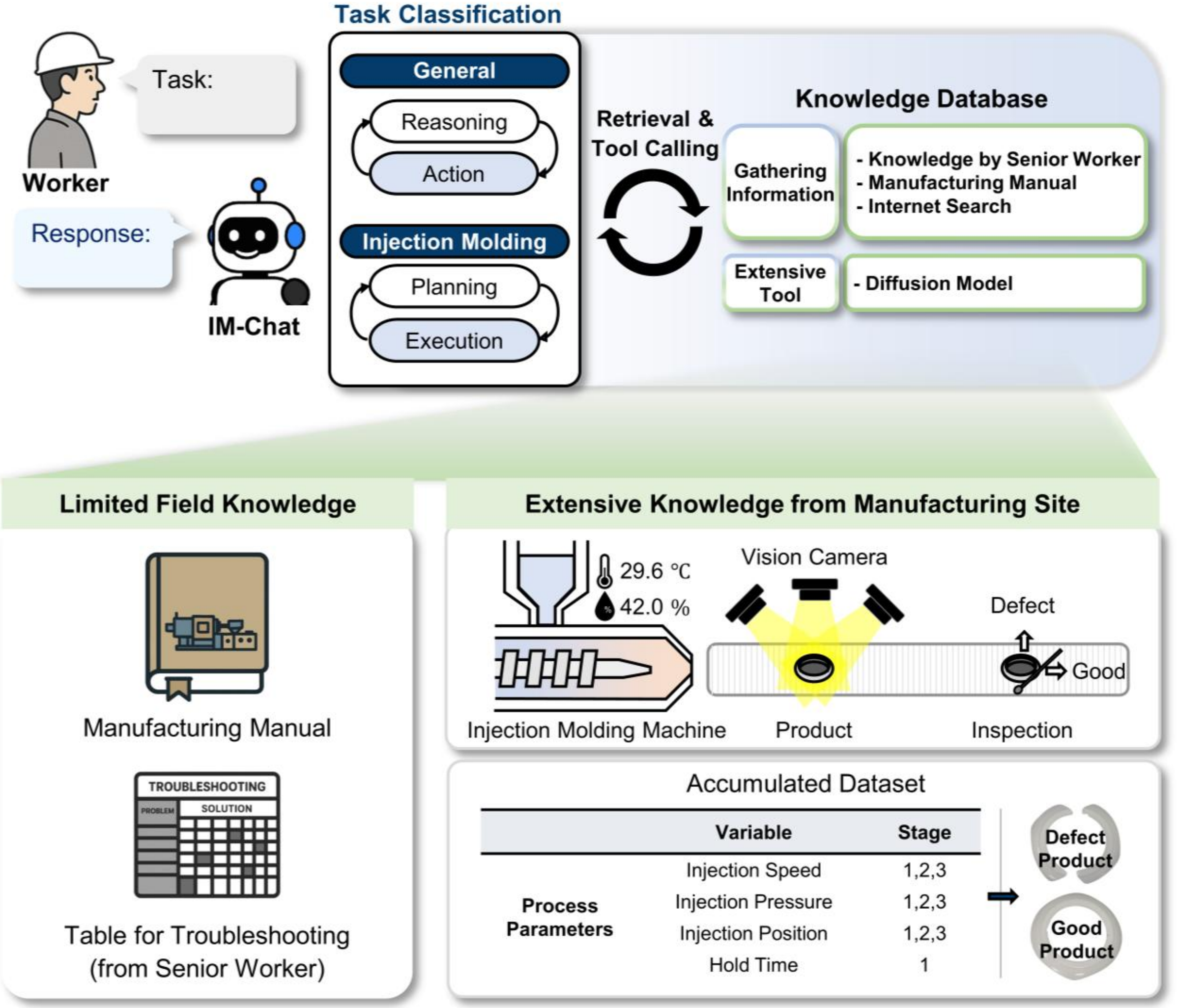

**Figure 1:** Schematic overview of the IM-Chat framework. Tasks are categorized as either general or injection molding–specific and are processed using ReAct or plan-and-execute workflows. The system integrates limited documented knowledge (e.g., manuals, troubleshooting tables) with extensive domain knowledge modeled by diffusion-based generative models trained on manufacturing field data.

## 1.2 Related Works

### 1.2.1 Knowledge Transfer in Manufacturing

Extensive research in the injection molding domain has accumulated substantial knowledge relevant to both process innovation and knowledge transfer. Prior studies have investigated rapid tooling methods, polymeric and composite mold materials, defect mechanisms, quality assurance, and recyclability, while others have provided reviews, explored sustainable manufacturing, or introduced multifunctional and additively manufactured tools [36–53]. Additional cross-disciplinary works on heat treatment, polymer flammability, composite behavior, and deformation methods further intersect with injection molding and enrich the accumulated expertise in manufacturing knowledge transfer [54–59].

Knowledge transfer approaches and expert systems have significantly contributed to advancements in production efficiency, knowledge management, and process optimization in the manufacturing and injection molding industries. Numerous studies have explored integrating knowledge-based methodologies with advanced computational tools to capture the expertise of skilled operators and effectively transfer it for broader application.

Early research established foundational frameworks by integrating group technology and schedule management to optimize production processes [60,61]. Subsequent studies further expanded these approaches by incorporating domain expertise into automated systems, reducing reliance on extensive datasets, streamlining parameter optimization, and minimizing operational costs [62]. In manufacturing operations, rule-based and expert systems have been widely applied to enhance process efficiency across various domains [63,64].

To improve knowledge transfer, frameworks have been developed to transform tacit knowledge into explicit knowledge, addressing uncertainties in production evaluation [65]. This is exemplified in systems that integrate expert knowledge into machine learning models for accurate quality predictions, even with limited data availability [66]. Additionally, case-based reasoning (CBR) has been applied extensively in injection molding manufacturing processes to assist in problem-solving and defect analysis [67].

Advancements in knowledge representation have brought significant improvements to the injection molding industry. The use of knowledge graphs has improved the structuring and accessibility of domain-specific information [68–70], while retrieval systems based on transformer architectures (e.g., BERT) have enhanced the efficiency and precision of knowledge queries [71,72]. Similarly, frameworks such as blackboard-based expert systems and interactive knowledge-based CAD systems have demonstrated effectiveness in areas such as mold design, defect analysis, and process planning by integrating both static rules and real-time operational data [73,74].

Other research has targeted manufacturability assessment and early-stage design optimization. Hierarchical knowledge frameworks and feature representations have improved manufacturability evaluations, while structured knowledge systems have helped mitigate risks and enhance design consistency in mold manufacturing [75,76]. For instance, interactive CAD systems and integrated knowledge management platforms have been instrumental in navigating mold design processes and managing historical design knowledge [77,78]. These systems are particularly valuable for small enterprises, enabling scalable and efficient knowledge transfer [79].

In parallel, expert systems have been applied to production planning and scheduling tasks. Approaches such as fuzzy decision models and multi-objective optimization frameworks have enabled improved resource allocation and operational flexibility in complex manufacturing scenarios [80,81]. More recent studies have introduced knowledge-based synthesis systems and preference-guided optimization strategies to enhance the adaptability of process modeling and configuration [82,83].

Despite these advancements, challenges persist in leveraging tacit knowledge and adapting to dynamic industrial conditions. Current systems largely focus on explicit knowledge, often neglecting the intuitive insights that arise from the experience of skilled workers. Additionally, many frameworks lack the flexibility to adapt to real-time changes in operational environments, limiting their scalability and practical application. Furthermore, significant technical expertise is often required to deploy and use these systems, creating barriers for less-skilled workers and restricting accessibility [84,85]. However, in the injection molding domain, research on multi-agent LLM-based approaches remains extremely limited, and few studies have investigated knowledge transfer frameworks that combine RAG with tool-calling mechanisms. In particular, no prior work has integrated a generative deep learning model, such as a diffusion model, as a callable tool to support quantitative decision-making within manufacturing workflows. Accordingly, this study introduces IM-Chat, which aims to address these limitations and advance knowledge transfer in injection molding.

### 1.2.2 LLM Application in Manufacturing

The integration of LLMs into manufacturing is accelerating the shift from Industry 4.0 to Industry 5.0, emphasizing human-centric innovation, enhanced automation, and adaptive intelligence [86–92]. Unlike traditional AI systems that primarily rely on structured datasets and predefined rules, LLMs possess reasoning capabilities, allowing them to process more flexible inputs and enhancing adaptive decision-making. Furthermore, techniques such as strategic prompting, RAG, tool-calling agents, and multi-agent systems further refine their accuracy and responsiveness.

Leveraging these advantages, recent research has increasingly explored their application in predictive maintenance, human-machine interaction, smart automation, process planning, and additive manufacturing, driving the development of more adaptive and efficient manufacturing environments [93]. The most straightforward way to integrate LLMs into manufacturing is through direct prompting, where users interact with a pre-trained model using carefully designed queries [94–96]. For instance, pre-trained LLMs alone have been successfully applied to automate the extraction and classification of knowledge from unstructured text, reducing the need for manual preprocessing in data analytics and machine learning workflows [94]. Furthermore, they have demonstrated effectiveness in generating context-aware text, which has been leveraged for various tasks such as creating, modifying, and refining code, enhancing automation in manufacturing processes [95].

While prompting offers utility in natural language tasks, its reliance on the model's pre-trained knowledge limits its ability to incorporate real-time data or specialized domain information, often leading to hallucinated or inaccurate outputs. To address this, several studies have fine-tuned LLMs using supplementary domain data such as machine documentation, fault logs, and taxonomies to improve their industrial relevance and task alignment [97–101]. In one example, fine-tuning enabled the model to infer the relationship between process parameters and product quality outcomes [102]. Alternatively, RAG provides a scalable means of enhancing factual grounding without the computational burden of fine-tuning. By retrieving relevant external knowledge and embedding it into the prompt, RAG enhances context alignment and output accuracy [97,100,103,104].

Beyond RAG, which solely integrates external information for answering queries, LLM agents can also interface with various tools, including sensors, control programs, robotic systems, and machine operators, to execute appropriate actions. Their ability to reason and take actions has played a crucial role in advancing the automation of manufacturing systems, spanning design, control, and production. For instance, an LLM agent integrated with specialized tools and domain-specific knowledge can enhance process planning for composite structures by optimizing decision-making and workflow efficiency [105]. As manufacturing applications advance toward higher levels of autonomy, multiple LLM-based agents collaborate to manage complex tasks. Several studies have explored such multi-agent frameworks, where distinct roles are assigned to each LLM agent, facilitating seamless communication, information exchange, and coordinated task execution [106–112].

# 2 METHODS

This section presents the architecture and implementation details of the proposed knowledge transfer system, IM-Chat, developed for the injection molding domain. We begin with an overview of the framework and its operational flow, describing how various components, including agents, tools, and knowledge sources, collaborate to process user queries (**Method 2.1**). We then outline two representative knowledge transfer scenarios: (i) tasks solvable through limited domain knowledge extracted from manuals and troubleshooting tables, supported via RAG (**Method 2.2**); and (ii) tasks requiring extensive field knowledge modeled from real-world production data, handled using data-driven generative models integrated into the agent workflow (**Method 2.3**).

## 2.1 IM-Chat Workflow

IM-Chat processes worker input tasks through three sequential stages: (i) input formatting, (ii) task solving, and (iii) output formatting (**Figure 2**). Upon receiving a task, the system first reformulates the input using context from recent conversation history to ensure clarity and minimize redundant LLM inference. This is handled by the Task Formatter, an LLM module that summarizes and restructures the concatenated user input and prior dialogue into a predefined query format. To support multilingual users, the formatted input is translated into English, and both the translated version and the original language metadata are retained for use during the final output stage.

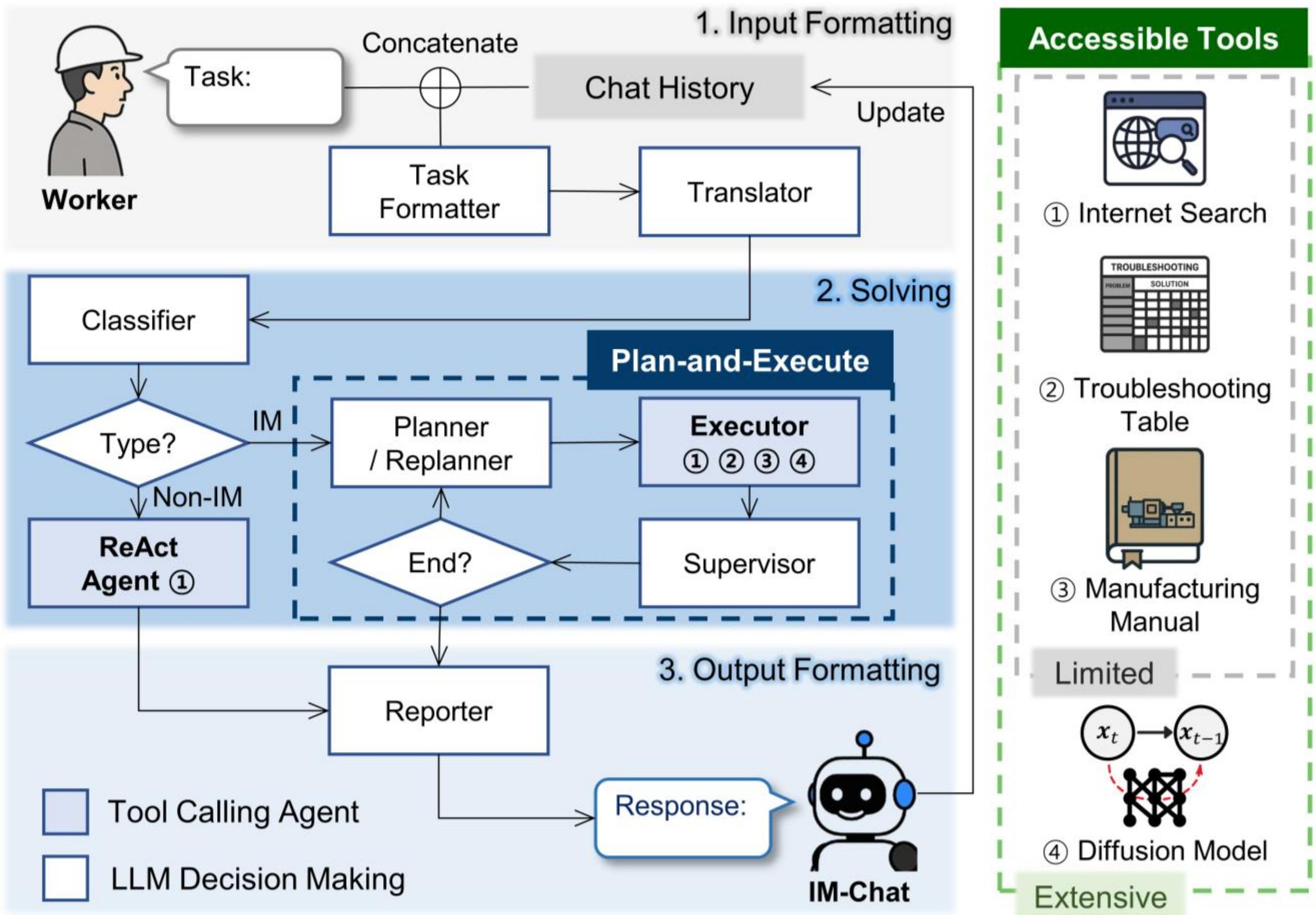

**Figure 2:** Overview of the IM-Chat framework for knowledge transfer in injection molding. The system consists of input formatting, task solving (via ReAct or plan-and-execute), and output formatting. Four tools (search, troubleshooting table, manual retriever, and diffusion model) are used during execution to generate responses.

Following input formatting, a *Classifier* module determines whether the query pertains to injection molding or not. This classification is necessary to preserve the generality of the framework, as using tools and prompts tailored for injection molding could otherwise constrain its broader applicability. It also helps minimize operational costs by preventing unnecessary invocation of specialized modules for simple, non-injection-related queries. In addition, it enables more specialized and carefully designed workflows for handling injection molding-specific tasks. The *Classifier* module, guided by a system prompt, is designed to output a parsed string, either "*injection*" or "*no_injection*", based on an overall assessment of the formatted input, which includes both the user's task and the preceding chat history, in conjunction with the capabilities of the available tools within the framework.

For general queries unrelated to injection molding, a *ReAct Agent* is employed [31], which operates solely with an internet search tool to retrieve recent information necessary for addressing the query. The use of a *ReAct Agent* allows the system to combine information retrieval and reasoning, enabling flexible handling of a wide range of general queries. By limiting the tools to internet search for non-injection-related tasks, the framework avoids unnecessary invocation of tools specialized for injection-related tasks, thereby reducing operational costs and maintaining system robustness. This design choice ensures that IM-Chat remains adaptable and efficient even when addressing queries outside of its core manufacturing focus.

For injection molding-related queries, the task is handled through a carefully designed planning and execution workflow. Upon receiving the formatted task, a *Planner* module decomposes the task into multiple subtasks. Each subtask is sequentially addressed, starting with the first subtask at the top of the plan. Specifically, the *Executor* module executes only the foremost subtask provided by the *Planner*, ensuring that the system maintains tight control over the process of solving the task. After execution, a *Supervisor* module evaluates whether the combined results sufficiently answer the original query. If the results are deemed satisfactory, the system proceeds to the output formatting phase. Conversely, if the results are insufficient, the *Replanner* redefines and adjusts the plan, triggering another cycle of execution.

During the execution phase, IM-Chat utilizes four external tools:

(i) an internet searcher,

(ii) a troubleshooting table retriever,

(iii) a manufacturing manual retriever, and

(iv) a diffusion model trained on real-world injection molding machine data

The first three tools provide limited but interpretable domain knowledge via a RAG strategy (**Method 2.2**). These tools, such as the troubleshooting table and the manufacturing manual, are not limited to a specific product and can be applied across a wide range of molding products regardless of their type, since they provide general guidance on how process parameters should be adjusted in response to different defect types. The diffusion model, by contrast, offers quantitative recommendations for process parameters under given environmental conditions (**Method 2.3**). Since the diffusion model is trained as a surrogate on production data from a specific product, its applicability is tied to the availability of production data. When such data are accessible, the model can serve as part of an extensive domain knowledge-based system that enables knowledge transfer for process parameters under given environmental conditions. To extend this capability beyond individual cases, the IM-Chat framework adopts a modular design that allows surrogate models to be trained separately for different products and flexibly replaced within the system. This modular structure ensures that the framework can accommodate diverse molding products while maintaining the consistency and scalability of its knowledge transfer functions.

After the task-solving phase is completed, the system proceeds to the response formatting stage, managed by a *Reporter* module. For non-injection molding-related tasks, the *Reporter* summarizes the response based on the original input and the generated answer. In the case of injection molding related tasks, the *Reporter* reviews both the original input and the temporary agent history accumulated during the processing of a single query, including the planning steps, tool calls, and execution results, to generate a concise and contextually accurate summary. The finalized response is then translated into the language originally used by the worker, ensuring seamless communication. Finally, the pair of input and response is appended to the global chat history, enabling the system to maintain continuity across interactions and support future reasoning processes.

The IM-Chat framework was implemented using the *LangChain* and *LangGraph* libraries, which provided the necessary infrastructure for orchestrating workflows involving multiple agents and reasoning processes using external tools. Detailed descriptions of each LLM module's prompt design is provided in the **Supplementary Note 1**.

## 2.2 LLM-based Knowledge Transfer System for Limited Domain Knowledge

### 2.2.1 Knowledge Extraction

To develop an effective knowledge transfer system powered by the LLM, it is essential to collect and structure the necessary production knowledge systematically. In this study, we gathered and extracted knowledge from various sources, including interviews with injection molding experts, troubleshooting charts standardized by the industry, and technical resources. Through expert interviews, we collected practical knowledge on how to adjust process parameters for specific defect types and prioritized these adjustments based on their effectiveness. Additionally, structured troubleshooting guidelines were referenced to address common defects such as short shots, weld lines, and warping, thereby elucidating the correlations between those defects and the associated process parameters [113]. Furthermore, technical books and reference materials on plastic injection molding enriched the knowledge base by validating expert opinions and troubleshooting data, offering a broader and more comprehensive context [114–117]. By integrating these insights, we identified the relationships between specific defect types such as burrs, short shots, and warping and process parameters like injection speed, injection pressure, and hold time. This allowed us to systematically prioritize corrective actions to produce products without defects. The compiled knowledge is summarized in a process adjustment table (**Figure 3a**), where each parameter is marked with a '+' or '–' sign to indicate whether it should be increased or decreased in response to a specific defect. In addition, the relative importance of each parameter is indicated by a priority value, with values closer to 1 denoting higher importance.

The operation manual for the injection molding machine was also considered a critical source of knowledge. For this study, we utilized the operation manual of a specific injection molding machine, available in PDF format, several pages of which are shown in **Figure 3b**. The manual spans a total of 227 pages and covers a wide range of operational and technical topics. It provided essential guidance on material-specific process parameter ranges, pre-operation checkpoints, and regular inspection schedules, as well as further operational details that support comprehensive field implementation. For example, the manual outlined the appropriate ranges for injection pressure, barrel temperature, and hold time for materials like ABS and PP, enabling operators to adjust process parameters effectively based on resin characteristics. Additionally, the manual included detailed preoperative checkpoints, such as inspecting lubrication systems, hydraulic components, and mold alignment, to prevent machine malfunctions and ensure stable operation. The regular inspection routines, categorized by weekly, monthly, quarterly, semi-annual, and annual schedules, specified tasks such as motor temperature checks, heat band cleaning, and hydraulic oil replacement to maintain optimal machine performance.

By combining the insights from expert interviews, troubleshooting charts, and machine manuals, this study developed a robust and comprehensive knowledge base. As a foundational component of the proposed LLM-based knowledge transfer system, this knowledge base is not limited to a specific product but can be applied across diverse molding products regardless of their type. It provides a broad range of practical guidance, such as how to adjust process parameters in response to different defect types, as well as material-specific parameter ranges and operational practices, thereby supporting adaptive and efficient decision-making across a variety of production scenarios.

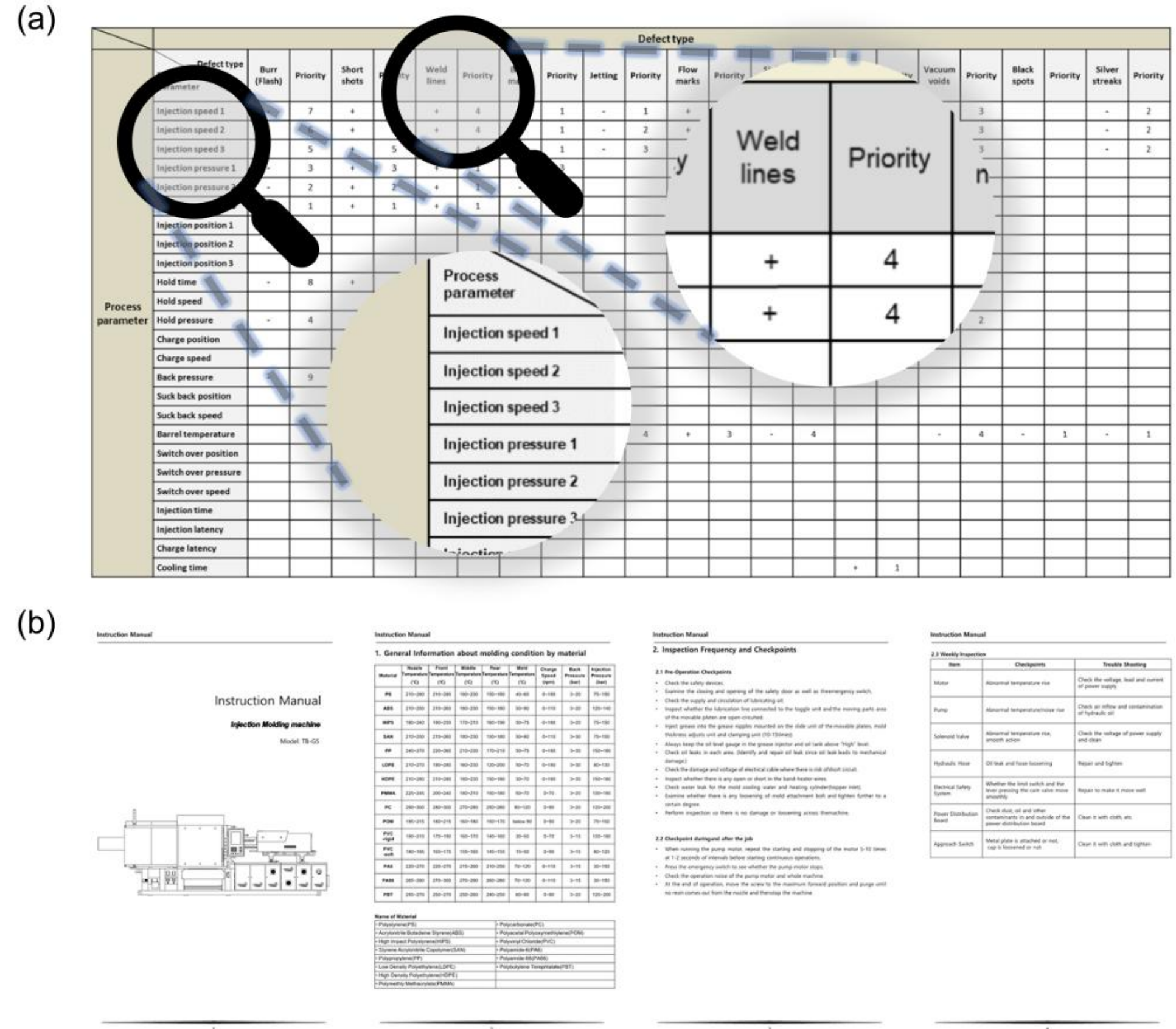

**Figure 3:** The documented knowledge base for IM-Chat. (a) The process adjustment table based on defect types. (b) The example pages from a 227-page machine operation manual.

### 2.2.2 Retrieval Augmented Generation (RAG) for IM-Chat

As described earlier, when an injection molding-related task is identified, IM-Chat does not rely solely on LLM reasoning. Instead, it aims to generate factually grounded response by retrieving relevant. For the documented knowledge type, IM-Chat employs a RAG approach to address input subtasks. Specifically, upon receiving the first subtask in the plan generated by the *Planner*, the *Executor* module dynamically selects the appropriate knowledge source, including a table retriever, manual retriever, or internet searcher, depending on the subtask requirements. After knowledge retrieval, the output is passed to the *Supervisor* module, which assesses whether the retrieved content sufficiently addresses the subtask requirements. Based on this assessment, the system either reports the response as final or returns to the *Replanner* to revise the plan and re-execute the subtask (**Figure 4a**). To ensure efficient retrieval and accurate grounding, the system applies distinct RAG configurations tailored to each type of knowledge source.

To begin with, the knowledge retrieval process for the troubleshooting table proceeds as follows. The table, stored in CSV format, is initially loaded and partitioned into chunks, where each chunk corresponds to a specific defect type or priority label. Each chunk contains detailed descriptions of recommended process parameter adjustments to address the corresponding defect, based on senior operators' experiential knowledge. Each chunk is then embedded into a vector space using OpenAI's embedding models, with one embedding vector representing one chunk. When a subtask query is issued, a similarity search is performed against the vector database using cosine similarity to retrieve the top two most relevant chunks. The retrieved information is subsequently summarized and reformulated, and the resulting augmented output, which consists of the input subtask query paired with the summarized retrieved information, is passed to the Supervisor for evaluation in the context of solving the original input task. This retrieval and augmentation procedure is illustrated in **Figure 4b**.

For the manufacturing manual, a structured retrieval pipeline was constructed to handle the large volume of information contained within the document. The manual, consisting of 227 pages in PDF format, was first processed using a vision-language model (GPT-4o) from the *LlamaParse* library. Each page was parsed into a structured JSON object, with the extracted content stored in Markdown format within each entry. This process enabled simple textual descriptions to be generated for embedded images, while tables were cleanly extracted and represented in a plain-text format with consistent alignment. Following this, each page-level chunk was embedded into a vector space using OpenAI's embedding, with one embedding vector representing one page. When a subtask query is issued, an initial similarity search is conducted using cosine similarity to retrieve the top 20 candidate chunks. From these candidates, 7 chunks are selected using the Maximal Marginal Relevance (MMR) method, which balances relevance to the query and diversity among the selected chunks. The MMR objective can be mathematically defined as:

$$\mathrm{MMR}(D_i) = \underset{D_i \in R \setminus S}{\operatorname{argmax}} \left[ \lambda \cdot sim(D_i, Q) - (1 - \lambda) \cdot \max_{D_j \in S} sim\left(D_i, D_j\right) \right] \tag{1}$$

where $D_i$ is the embedding of a candidate chunk under consideration, $R$ is the set of retrieved embeddings, $S$ is the set of already selected embedding vectors, $Q$ is the query embedding, and $\lambda$ controls the trade-off between relevance and diversity. Here, $sim(\cdot,\cdot)$ denotes the cosine similarity between two embedding vectors [118]. The selected chunks are then summarized and reformulated in combination with the subtask query, and the resulting augmented output is passed to the *Supervisor* for evaluation. This retrieval procedure is visually summarized in **Figure 4b**.

Finally, for cases where the required information is not available from the internal troubleshooting table or the manufacturing manuals, an external internet search is performed. The *Tavily* API is adopted to retrieve external information relevant to the subtask query. The retrieved knowledge is processed similarly through summarization and augmentation steps before being passed to the Supervisor for evaluation. In situations where external retrieval is unnecessary or impractical, the Executor may opt to solve the subtask purely through internal LLM inference without invoking any tools. The specific prompts used for each retrieval tool are provided in **Supplementary Note 2**.

## 2.3 LLM-based Knowledge Transfer System for Extensive Domain Knowledge

### 2.3.1 Diffusion Model

Knowledge in the injection molding process extends beyond general insights obtained from expert interviews or technical resources on adjusting process parameters for specific defect types. While these sources offer a strong foundation, incorporating extensive production data significantly broadens the scope of field knowledge. With sufficient production data, deeper knowledge extraction and application become possible, transforming raw data into a valuable resource for improving manufacturing processes. For example, real-world production data enables more accurate factory simulations, allowing manufacturers to replicate actual operating conditions and dynamically optimize process parameters. Moreover, such data supports the development of the knowledge transfer system capable of providing precise solutions for determining optimal processing conditions, surpassing traditional heuristic approaches. By integrating knowledge derived from both experts and data, a more adaptive and precise system can be established, ultimately enhancing production efficiency and quality control in complex manufacturing environments.

Among various forms of domain knowledge, production data plays a critical role in advancing knowledge transfer systems that integrate expert insight with data-driven approaches. To explore this potential, we developed an external recommendation tool based on diffusion model to suggest optimal process parameters for defect-free production. Because the diffusion model is trained as a surrogate on data from a specific product, its scope of application depends on the availability of such production data. To enhance adaptability, the IM-Chat framework was designed with a modular structure that allows different surrogate models to be trained for various products and flexibly incorporated into the system. This modular design ensures that the framework can accommodate diverse molding products while maintaining consistency and scalability in its knowledge transfer functions. Based on this design, we trained the diffusion model using structured production data that included environmental conditions (factory and machine temperature and humidity), process parameters such as injection speed and pressure, and the corresponding product quality outcomes. To clarify how these data were obtained and structured, and to demonstrate their relevance to real-world manufacturing, the data collection process is described below.

To develop the diffusion model based external recommendation tool, real production data were collected from a factory that manufactures cosmetic container caps made of ABS resin. This product was selected because it is one of the most widely produced items in the plastic industry. Its circular shape and the ABS resin material used are both broadly representative in injection molding practice. Over a six-month period, data were systematically gathered using a dedicated testbed. The dataset covered three categories of data: environmental data, product quality data, and process parameter data. Environmental data, including machine and factory temperature and humidity, were continuously recorded every minute using attached sensors. Product quality data were automatically inspected using a vision-based system that evaluated both dimensional accuracy and surface defects. Dimensional accuracy was assessed by measuring whether the inner diameter fell within the tolerance range of 48.6–48.9 mm. Surface defects included burrs, short shots, and weld lines, which were detected by the automated inspection system.

Process parameter data were obtained through a central monitoring system (CMS) that logged 43 operator-input production settings across all production runs. Injection molding includes a large number of process parameters such as injection speed, injection pressure, injection position, holding time, mold temperature, and melt temperature. Considering all parameters simultaneously and collecting sufficient experimental data for them is practically infeasible. In real mass production environments, operators selectively manage a small set of variables that are most relevant to quality control rather than controlling every parameter. To identify these parameters, historical production data from the testbed were analyzed, and this confirmed that 10 process parameters were consistently adjusted by operators as the primary controllable factors. These include three-stage injection speeds, three-stage injection pressures, three-stage injection positions, and hold time. Parameters such as mold temperature and melt temperature are undoubtedly important, but they are automatically regulated by the machine to remain within allowable ranges specified for each resin and product. Since they do not act as uncontrolled disturbances in practice, they were not included as input variables for training the diffusion model. The relevance of the selected variables was further validated by shapley additive explanations (SHAP) analysis, which showed a strong correlation between these operator-adjusted parameters and product quality outcomes.

To ensure systematic data collection and balanced experimental coverage of the selected 10 process parameters, an orthogonal experimental design was introduced. An orthogonal experimental design (L81) was applied to these parameters, while environmental variables varied naturally under factory conditions. In total, 2,794 experimental data points were collected. Each data point mapped 10 process parameters, 4 environmental variables, and a binary quality label indicating whether the product was good or defective. Among them, 2,571 corresponded to defect-free products and 223 to defective products. The overall defect rate of approximately 5 percent aligns with real-world production settings. Although the dataset exhibits class imbalance, the inclusion of diverse environmental variations and multiple defect types enhances its representativeness. Further details of the data collection procedure and SHAP analysis results are available in Kim et al [6].

To implement this system, we constructed an external tool based on the Classifier-Free Guidance Diffusion Model (CFGDM) [119], a type of score-based generative model known for its training stability and sample quality. The model takes five input features: product class (good or defective), factory temperature, factory humidity, machine temperature, and machine humidity. Based on these inputs, it generates ten output parameters that guide defect-free production, including three-stage injection speeds, three-stage injection pressures, three-stage injection positions, and hold time. A trained diffusion model was evaluated under previously unseen environmental conditions, where 64,000 process parameters intended to produce non-defective products and 64,000 process parameters intended to produce defective products were generated, respectively. These generated samples were validated using a high performance classification model, referred to as the surrogate model, trained on the same dataset. The surrogate model achieved a true positive rate (the proportion of actual defective products correctly predicted as defective) of 0.927±0.066, and a true negative rate (the proportion of actual non defective products correctly predicted as non-defective) of 0.995±0.006. Functioning as a surrogate for the injection molding machine, the classification model enabled virtual experiments by verifying whether the generated process variables aligned with their given labels. As a result, among the 64,000 variables intended to yield non defective products, 799 cases (1.248%) were incorrectly generated, while among the 64,000 variables intended to yield defective products, 1,288 cases (2.013%) were misgenerated, thereby confirming the high reliability of the CFGDM. In addition, CFGDM was validated in a real production environment, where three separate experiments were conducted at different time points to evaluate performance under varying environmental conditions. In each experiment, five process parameters intended to produce non-defective products were generated and subsequently input into the injection molding machine to produce physical products. Among the 15 products obtained across the three experiments, 13 were successfully produced as non-defective. Notably, in all instances where the surrogate model predicted a product to be non-defective with a probability of at least 0.99, the experimental results were consistently non-defective. Furthermore, the inference time on an NVIDIA RTX 3090 GPU was approximately 40 seconds for generating 64 process parameters. These results highlight the strong potential of the proposed approach for real-world application. Details of the diffusion model architecture, training procedure, and evaluation results are provided in **Supplementary Note 3**, with further technical explanations available in Kim et al [6].

### 2.3.2 Diffusion Model Tool-Calling for Recommending Process Parameters

To further clarify the respective roles of the diffusion model and the LLM within this pipeline, the diffusion model serves as a quantitative generator that produces candidate process parameters under given environmental conditions, while the LLM interprets and contextualizes these outputs by integrating them with domain knowledge to provide actionable recommendations. In addition, the LLM is responsible for input formatting, converting natural language instructions into structured vector representations compatible with the diffusion model, thereby bridging human language and numerical input spaces. This design allows users to utilize the diffusion model through natural language, even without expertise in its underlying algorithms or configuration. The LLM thus acts as both an interface and an interpreter that manages the model's inputs and outputs, enabling seamless interaction with complex generative models.

Therefore, to leverage the trained diffusion model for recommending process parameters, it is essential to provide structured and well-defined input conditions based on the worker's input natural language query, as the model requires inputs in a specified vector format. To ensure this, an *Diffusion Input Formatter* module is placed before the diffusion model to standardize and organize incoming queries. If the initial input from the worker lacks sufficient detail, the system prompts the worker to supplement or clarify the information to meet the required specifications. Once properly formatted, the input is passed to the diffusion model, which generates 64 candidate outputs representing potential process parameter sets tailored to the specified conditions.

Following candidate generation, a CatBoost-based surrogate model is employed to filter and rank the outputs by predicting the likelihood of defect-free production based on the generated process parameters. Each candidate is scored according to its predicted probability of yielding a good (non-defective) outcome [120]. The candidate with the highest predicted success probability is selected, and its corresponding process parameters are reformulated into a structured textual output. This refined result is then integrated into the overall workflow and passed to the *Supervisor* for evaluation alongside other task-solving outputs. This pipeline is illustrated in **Figure 4c**. Representative system prompts for both the input formatter and the output parsing format are provided in the **Supplementary Note 4**.

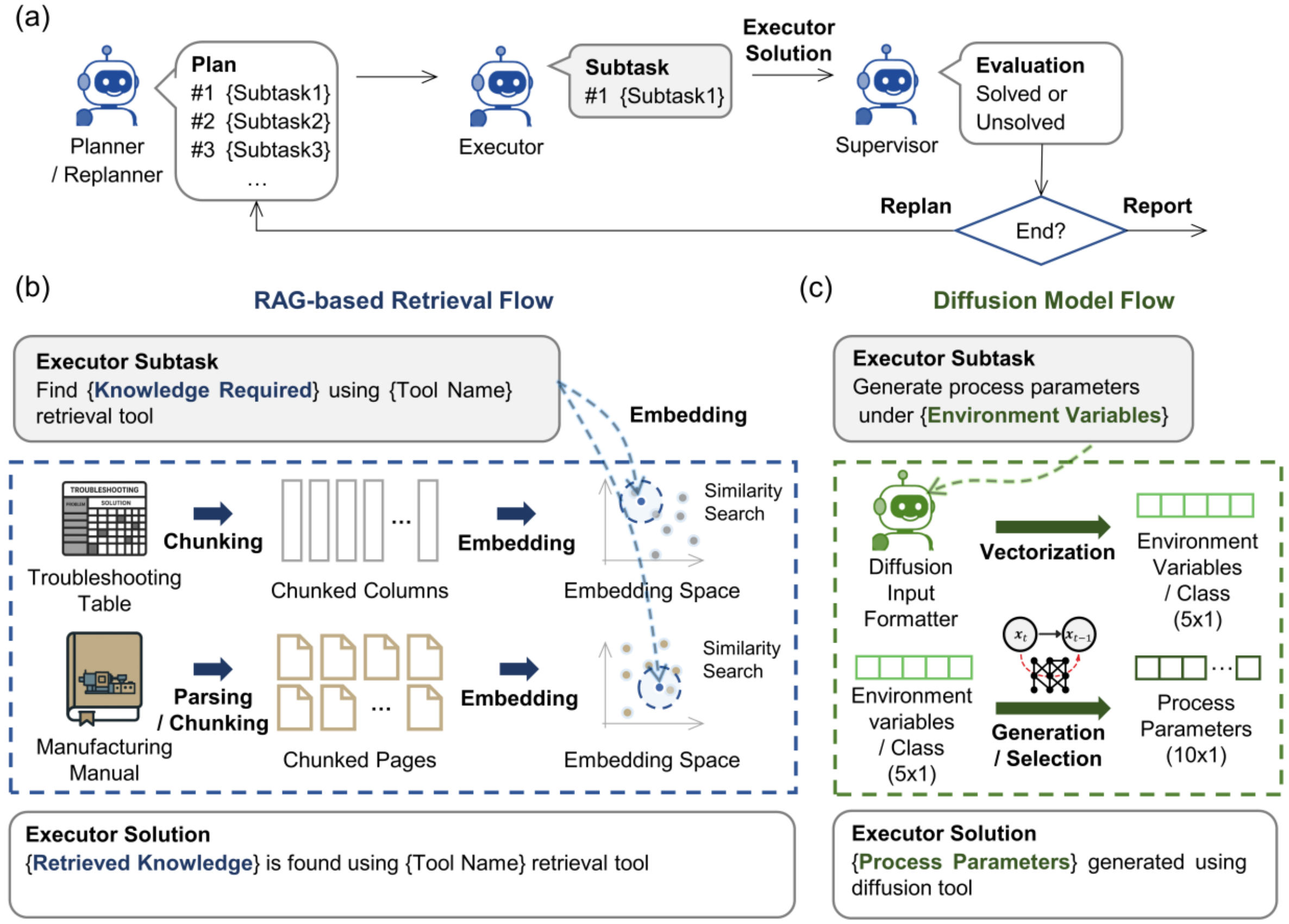

**Figure 4:** Execution framework of IM-Chat for injection molding task resolution. (a) Plan–execute–replan loop overview. (b) RAG-based knowledge retrieval flow using the troubleshooting table and manufacturing manual. (c) Diffusion model flow for process parameter generation from textual environmental inputs using a diffusion model.

# 3 RESULTS

## 3.1 Validation of IM-Chat with Single-Tool Task

Since IM-Chat incorporates a translator module that standardizes all user inputs into English before task execution, it inherently supports multilingual interactions. As an initial check, we validated the system using a small set of representative queries in four languages (Korean, English, Vietnamese, and Thai), considering the multilingual environment commonly observed in the injection molding industry in South Korea. These results, summarized in **Supplementary Note 5**, indicate that IM-Chat accurately processed the translated queries and returned contextually appropriate responses. Therefore, in the subsequent evaluations, we focused on English-language queries, as all user inputs are translated into English by default before processing.

Building on this, we conducted a comprehensive evaluation to benchmark the system's reasoning and tool selection capabilities using a curated set of 100 English-language queries. Specifically, in this section, we tested IM-Chat's performance on 100 single tool tasks, where each query was designed to invoke one of four specific tools: troubleshooting table retrieval, manufacturing manual retrieval, diffusion model generation, or ordinary reasoning tasks that may or may not involve internet search. For example, a single-tool task that requires the troubleshooting table retrieval tool is: "How should injection speed be adjusted to reduce burr defects?" This type of query was intentionally designed so that a correct response can only be generated by invoking the troubleshooting table, which contains process parameter adjustment strategies for addressing burr defects. In this manner, for each category, 25 representative queries were prepared, resulting in a dataset of 100 question–answer pairs (i.e., chatting logs).

To assess the quality of the previously generated question and answer pairs, we developed a customized evaluation metric tailored to the specific needs of the injection molding domain. Conventional automatic metrics widely used in natural language processing, such as BLEU, ROUGE, perplexity, and semantic similarity [84,85], mainly focus on linguistic fluency or surface-level similarity. While these metrics are effective for general language tasks, they are not suitable for evaluating technical accuracy and domain relevance, which are essential in manufacturing tasks that require expert-level understanding. In addition, most queries in injection molding have clearly defined and objective answers, making the factual correctness and appropriateness of the retrieved information especially important. These considerations led us to design a domain-specific evaluation framework that better reflects the practical performance of IM-Chat.

Based on the customized rubric, human experts were asked to evaluate each response on a 10-point scale in terms of relevance and correctness. In addition to expert evaluation, an automated LLM-based assessment was also conducted, inspired by the recently proposed “LLM-as-a-judge” approach [121]. Specifically, GPT-4o was prompted with an evaluation instruction adapted to the domain (**Supplementary Note 6**) that guided the model to assess the responses with a focus on both correctness and relevance. This LLM-based assessment was conducted on the same 10-point scale to explore whether a language model could reliably approximate or potentially substitute for expert evaluation, by providing scalable benchmarking and independent judgments. The overall evaluation procedure is illustrated in **Figure 5a**.

The left panel of **Figure 5b** shows the evaluation scores assigned by human experts across four task categories and three IM-Chat variants (GPT-4o, GPT-4o-mini, and GPT-3.5-turbo). Overall, GPT-4o exhibited a consistent trend of achieving higher scores across most task types, with particularly strong performance on troubleshooting table and diffusion model tasks, indicating superior technical grounding and response integration capabilities. In contrast, GPT-3.5-turbo recorded the lowest scores in most categories, particularly struggling with retrieving information from manufacturing manuals and invoking the diffusion model, highlighting its limitations in handling domain-specific generation tasks. More specifically, GPT-3.5-turbo often failed to comprehend the context of lengthy documents or to extract correct values from tables embedded within manuals. It also exhibited a notably higher frequency of tool misselection during both the planning and execution stages.

The right panel of **Figure 5b** presents evaluation scores assigned by GPT-4o acting as an automated evaluator. Broadly, a similar pattern of performance degradation across task categories was observed. However, the LLM evaluator consistently assigned low scores to diffusion model tasks, which often involve interpreting numerical outputs typically encountered in real manufacturing environments. This suggests that when it comes to evaluating responses that reflect numerical reasoning encountered in the field, especially those produced by deep learning models trained on knowledge specific to the domain, evaluation by LLMs may fall short of accurate judgment. The LLM evaluator may be more sensitive to fluency or coherence at the surface level than to the domain correctness of the generated content.

To further assess the agreement between human and LLM evaluations, we computed Pearson correlation coefficients between the two scoring approaches across the three IM-Chat variants. The resulting correlations were −0.14 (GPT-4o), 0.25 (GPT-4o-mini), and 0.35 (GPT-3.5-turbo). These relatively low values indicate weak alignment between the LLM evaluations and expert human evaluations, particularly in the case of models that demonstrated superior performance. This misalignment highlights that while LLM-based assessment can provide scalable benchmarking, it cannot yet be relied upon as a full substitute for expert judgment, especially in industrial domains requiring precise technical evaluation, such as injection molding. These findings underscore the importance of human-in-the-loop validation when evaluating domain-specific chatbot systems.

**Figure 5c** illustrates the average response latency per query for each IM-Chat variant. GPT-4o demonstrated a favorable trade-off between latency (24.4 seconds) and accuracy, consistently outperforming the other models in correctness. In contrast, GPT-3.5-turbo achieved the fastest response time (17.2 seconds), but this speed often came at the expense of performance quality, as it tended to produce shallow reasoning and premature answers. Notably, GPT-4o-mini exhibited the highest latency (32.4 seconds), despite being a lighter model. This elevated latency was primarily caused by inefficiencies in its plan-and-execute loop, where suboptimal planning and unnecessary re-planning led to excessive iterations and indecisive behavior. These findings underscore that latency alone is insufficient to assess overall system effectiveness. Effective deployment requires not only fast inference, but also coherent reasoning strategies and well-coordinated agentic control.

**Figure 5d** presents the average cost per query for each IM-Chat variant. GPT-4o incurred the highest cost at $0.0190 per query – approximately 10 times more expensive than GPT-4o-mini ($0.0017) and about 7 times more than GPT-3.5-turbo ($0.0027). This reflects the premium pricing associated with GPT-4 class modes, which offer state-of-the-art performance but at significantly higher inference cost. These results emphasize that pricing considerations should be carefully weighed against performance and latency metrics when selecting LLM variants for real-world deployment.

In summary, the evaluation of IM-Chat across 100 curated single-tool tasks shows that GPT-4o achieved the highest accuracy and technical reliability but with considerably higher latency and cost. GPT-4o-mini and GPT-3.5-turbo produced faster and more affordable responses but with lower accuracy and less reliable tool selection. The results indicate clear trade-offs among accuracy, latency, and cost. In addition, the correlation analysis revealed weak alignment between human and LLM-based evaluations, particularly for tasks involving numerical reasoning. Overall, these findings describe the relative strengths and limitations of different IM-Chat variants in terms of accuracy, latency, cost, and evaluation consistency.

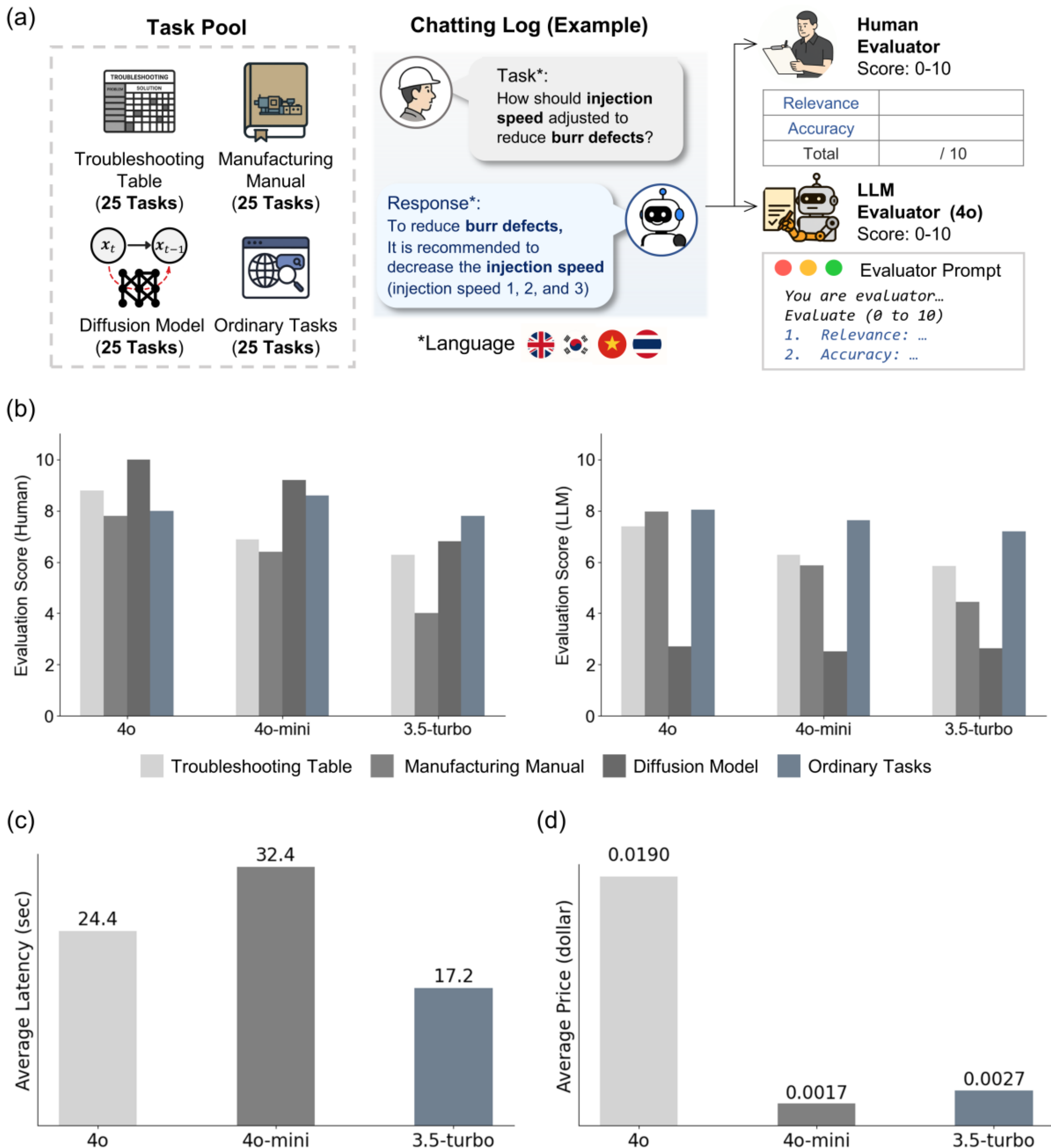

**Figure 5**: Evaluation of IM-Chat performance for single-tool tasks. (a) Overview of the evaluation framework including task types, evaluation setup, and scoring criteria. (b) Evaluation scores from human experts (left) and an LLM evaluator (right). GPT-4o achieved the highest performance, while GPT-3.5-turbo underperformed due to frequent tool misselection. (c) Average response latency per query, with GPT-3.5-turbo being fastest but least reliable. (d) Average query cost, with GPT-4o being most expensive due to deeper reasoning and higher token usage.

### 3.2 Validation of IM-Chat for Multi-Tool Tasks

To further assess the capabilities of IM-Chat in more complex and realistic industrial scenarios, we designed a set of hybrid tasks that combine limited knowledge retrievable from documentation or the internet with extensive domain knowledge embedded within a diffusion model trained on field data. **Figure 6a** illustrates the evaluation framework applied to these tasks. The task pool consists of three hybrid categories, each comprising 20 tasks: (1) Diffusion model + Troubleshooting table, (2) Diffusion model + Manual retrieval, and (3) Diffusion model + Internet search. Each task is specifically designed to require the combined use of both tools listed in its category, ensuring that IM-Chat must effectively coordinate multiple knowledge sources to generate a valid and complete response. For example, one representative query from the Diffusion model + Troubleshooting table category is: "When the Machine temperature is 22.0 °C, Machine humidity is 43.5%, Factory temperature is 26.0 °C, and Factory humidity is 34.0 °C, generate process parameters using the Diffusion model. Then, assuming a short shot defect occurs under these conditions, tell me how the Injection speed should be set based on the reference table." This query is intentionally structured to first require the generation of baseline process parameters using the diffusion model and then prompt a subsequent adjustment decision that can only be answered correctly by consulting the troubleshooting table, thereby ensuring that both tools should be effectively invoked to provide a valid response.

First, to verify that IM-Chat functions with multilingual inputs in multi-tool tasks, we provided representative examples of responses in Korean, English, Vietnamese, and Thai in **Supplementary Note 5**, following a procedure similar to that used for the single-tool tasks. Because each query is translated into English by the Translator module before being processed, the internal reasoning flow within IM-Chat is conducted entirely in English, and the final responses are translated back into the language originally used by the user. Hence, the subsequent evaluations were performed using English queries for consistency, similar to the single-tool evaluation.

Each pair of input query and generated response was evaluated using the same protocol described in the evaluation involving a single tool: human experts assessed the relevance and accuracy of the chatbot's response using a 10-point rubric, while GPT 4o also served as an automated evaluator using a prompt adapted to the domain.

**Figure 6b** compares evaluation scores for hybrid tasks involving both retrieval and reasoning based on diffusion models across three variants of IM-Chat. According to the human expert evaluation, GPT-4o significantly outperforms the other two models across all hybrid task categories. In particular, it exhibits outstanding performance in tasks combining diffusion models and troubleshooting (score = 9.2), while maintaining relevantly strong accuracy in hybrid tasks involving manuals and internet-based sources (scores = 7 and 5.4, respectively). This superior performance reflects not only the model's intrinsic reasoning capability but also the effective coordination among its internal components: the Planner, the Executor, and the Supervisor. These components work in concert to decompose complex queries that span multiple sources, select the appropriate tools (e.g., diffusion model or document retrievers), and ensure the consistency of the generated responses. As a result, GPT-4o demonstrates a robust ability to integrate numeric inference with structured retrieval or contextual understanding from static sources. In contrast, GPT-3.5-turbo and GPT-4o-mini show substantial performance degradation, especially in tasks requiring deeper coordination between retrieved knowledge and generated inferences.

In contrast to human evaluations, the assessments conducted by the LLM show compressed score distributions and weaker model differentiation. GPT-4o, which was clearly superior in human evaluations, received moderate but not outstanding scores from the LLM evaluator (all ~3–4). Meanwhile, GPT-3.5-turbo scores the highest among the three models, which contradicts the human assessments. This discrepancy suggests that the LLM evaluator may overestimate fluency or plausibility, particularly when responses appear reasonable at the surface level, even if they lack the necessary technical rigor. The evaluation prompt may still face challenges in accurately assessing reasoning that is both numerically grounded and derived from multiple information sources. In particular, integration of diffusion model outputs may not be easily validated by an external language model prompt without access to intermediate reasoning steps.

**Figure 6c** presents the average response latency per query for hybrid tasks involving both diffusion model inference and retrieval. GPT-4o recorded the highest latency (66.0 seconds), followed closely by GPT-4o-mini (63.2 seconds), while GPT-3.5-turbo responded substantially faster (33.3 seconds). This latency pattern primarily reflects the overhead of diffusion model execution, which takes approximately 37 seconds per inference and dominates the overall response time. The notably lower latency of GPT-3.5-turbo is partially due to its frequent failure to invoke the diffusion model when required, resulting in shorter but incomplete reasoning paths. This discrepancy explains its paradoxically fast response times despite lower task performance.

**Figure 6d** shows the average query cost for multi-tool tasks involving both retrieval and diffusion model inference. GPT-4o incurred a significantly higher cost ($0.0427 per query), more than twice the cost of its single-tool usage. This increase stems from longer reasoning traces, higher token usage, and multiple agent interactions required to coordinate complex tasks. In contrast, GPT-4o-mini and GPT-3.5-turbo maintained low costs (~$0.0027–0.0028), although the latter's cost remained artificially low due to frequent omission of diffusion model invocation, as observed in earlier latency and accuracy analyses.

In summary, the evaluation of IM-Chat on 60 hybrid multi-tool tasks showed that GPT-4o achieved the highest accuracy, particularly in tasks combining diffusion model inference with troubleshooting, while GPT-4o-mini and GPT-3.5-turbo exhibited reduced reliability in coordinating multiple sources. Human expert and LLM-based evaluations diverged, underscoring limitations of automated assessment. Latency was largely driven by diffusion model execution, with GPT-4o and GPT-4o-mini showing longer but complete reasoning paths, whereas GPT-3.5-turbo appeared faster mainly because it frequently skipped diffusion model invocation, leading to incomplete and inaccurate outputs. Cost analysis similarly highlighted GPT-4o's higher overhead, consistent with the single-tool evaluation. Compared with the single-tool evaluation, these hybrid tasks amplified the same trade-offs: superior accuracy with higher cost and latency versus efficiency at the expense of reliability.

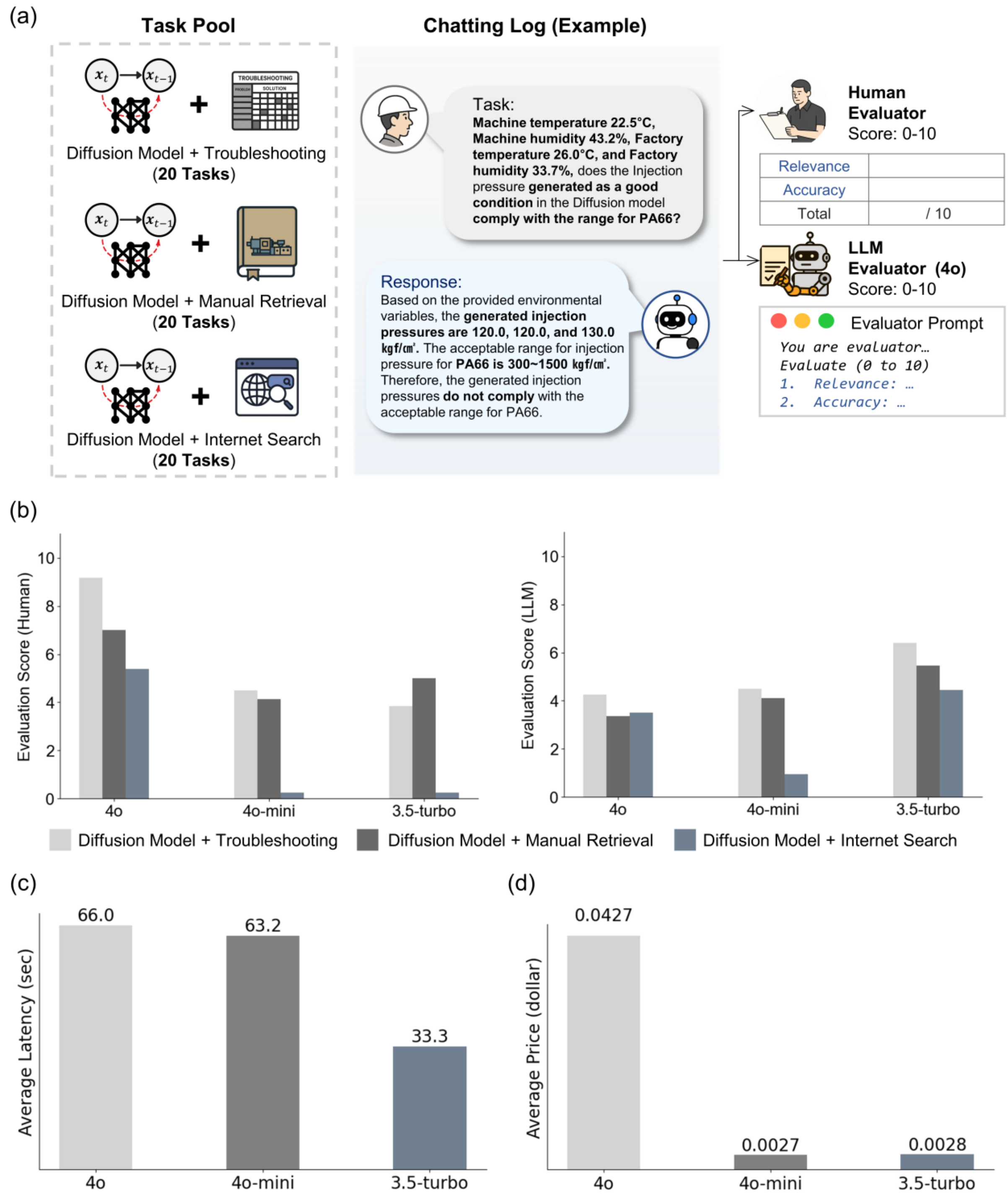

**Figure 6:** Evaluation of IM-Chat performance for hybrid (multi-tool) tasks. (a) Overview of the evaluation framework including task types, evaluation setup, and scoring criteria. (b) Evaluation scores from human experts (left) and an LLM evaluator (right). GPT-4o showed the best overall performance, while GPT-3.5-

turbo failed to invoke required tools, resulting in lower human scores. (c) Average response latency per query, with GPT-3.5-turbo being fastest but often incomplete in execution. (d) Average query cost, with GPT-4o being most expensive due to complex tool coordination and higher token usage.

### 3.3 Comparison with the Baseline

To evaluate the effectiveness of the proposed multi-agent approach for knowledge transfer in the injection molding domain, we developed a fine-tuned single-agent LLM for comparison. To emulate the training procedure of previous studies [99], we generated and curated conversational question–answer pairs representative of those typically asked by manufacturing workers, aiming for knowledge transfer-specific fine-tuning. These questions covered the same scope as the tools integrated within IM-Chat, including the troubleshooting table, manufacturing manual, process parameter proposal (diffusion model), and ordinary reasoning or searching tasks. The detailed procedure for training data generation is described in **Supplementary Note 7**. Both models were evaluated using the identical set of test queries employed in **Sections 3.1** (single-tool tasks) and **3.2** (multi-tool tasks). Finally, we compared the performance of the proposed IM-Chat (multi-agent LLM) with that of the fine-tuned single-agent LLM, both implemented using the GPT-4o model. The comparison results in terms of human-evaluated scores, response latency, and pricing are summarized in **Table 1**.

**Table 1:** Performance, latency, and cost comparison between the proposed IM-Chat multi-agent system (GPT-4o-based) and the fine-tuned single-agent LLM (GPT-4o-based) baseline across single- and multi-tool tasks.

| | | IM-Chat (Proposed) | | | Fine-tuned single LLM (Baseline) | | |
|---|---|---|---|---|---|---|---|
| | | Score | Latency (sec) | Pricing (dollar) | Score | Latency (sec) | Pricing (dollar) |
| Single-tool tasks | Troubleshooting table | 8.80 | 14.80 | 0.0263 | 7.00 | 1.15 | 0.0006 |
| | Manufacturing manual | 7.80 | 16.92 | 0.0207 | 5.40 | 1.40 | 0.0008 |
| | Diffusion model | 10.00 | 55.38 | 0.0209 | 2.00 | 1.67 | 0.0018 |
| | Ordinary tasks | 8.00 | 10.58 | 0.0081 | 7.60 | 1.14 | 0.0007 |
| | Total | 8.65 | 24.4 | 0.0190 | 5.50 | 1.34 | 0.0010 |
| Multi-tool tasks | Diffusion model + Troubleshooting | 9.18 | 63.42 | 0.0337 | 3.50 | 2.13 | 0.0029 |
| | Diffusion model + Manual | 7.00 | 66.87 | 0.0381 | 1.88 | 1.07 | 0.0014 |
| | Diffusion model + Ordinary | 5.38 | 67.58 | 0.0562 | 2.13 | 1.37 | 0.0138 |
| | Total | 7.19 | 66.0 | 0.0427 | 2.50 | 1.52 | 0.0060 |

The proposed multi-agent framework (IM-Chat) consistently outperformed the fine-tuned single-agent LLM in terms of response relevance and accuracy across all categories. In contrast, the fine-tuned single-agent model exhibited substantially lower latency and inference cost, since it relies on a single forward pass without tool coordination. Overall, this comparison reveals a clear trade-off between efficiency (latency and cost) and accuracy (response quality).

In terms of response quality, the fine-tuned model frequently produced hallucinated or inconsistent answers, particularly in tasks requiring numerical or data-driven inference. This limitation was most apparent in the diffusion model–related tasks, where the single-agent LLM achieved only a score of 2.00, compared to 10.00 by IM-Chat in single-tool tasks. Although both the diffusion model and the fine-tuned LLM were trained with the same dataset, these results demonstrate that training a diffusion model and invoking it through tool-calling yields significantly higher accuracy than directly fine-tuning the LLM alone.

Moreover, in multi-tool tasks, the fine-tuned LLM showed a pronounced decline in performance when it was required to coordinate partial information from multiple sources. This suggests that while fine-tuning can embed domain trends and factual knowledge, it does not inherently enable logical reasoning or coordination across heterogeneous information sources, which are capabilities naturally supported by the multi-agent design.

Enabling a fine-tuned model to effectively utilize information from multiple and heterogeneous sources would require constructing a large number of composite Q&A datasets that explicitly encode such interactions. This approach not only demands substantial manual effort but also limits scalability, as integrating new knowledge sources would necessitate additional data curation and retraining. In contrast, the modular structure of the multi-agent framework allows new tools or information sources to be incorporated seamlessly without retraining, highlighting its superior adaptability and scalability for dynamic industrial environments.

Regarding latency and cost, the fine-tuned single-agent LLM achieved much faster and cheaper responses (1.15–1.67 seconds, $0.0006–0.0018) compared to IM-Chat (10.58–55.38 seconds, $0.0081–0.0209) for single-tool tasks. This tendency was also observed in multi-tool tasks. Even when excluding cases that required invoking the diffusion model, the baseline system produced faster and more cost-efficient responses, since it generates an answer through a single inference step without additional tool calls. Therefore, for simple queries or short document-based questions, fine-tuning can be a practical choice. However, it should be noted that this approach requires an initial investment of $39.95 and 1 hour 43 minutes of training. Hence, despite the upfront cost and potential compromises in both accuracy and scalability, fine-tuning remains advantageous when low-latency and low-cost responses to relatively simple and well-defined queries are prioritized.

Overall, a clear trade-off emerges between the two paradigms. The proposed multi-agent approach is more suitable for scalable, reasoning-intensive applications, particularly those requiring multi-tool orchestration and quantitative inference, whereas the fine-tuned single-agent LLM is more appropriate for large-scale, low-cost knowledge retrieval or FAQ-style applications. Given that manufacturing knowledge transfer often involves qualitative judgments and quantitative reasoning (e.g., process parameter estimation via diffusion models), the integration of multi-agent architectures appears highly advantageous and increasingly necessary for achieving reliable and interpretable decision support in industrial environments.

## 4 DISCUSSION

This study presents a multi-agent LLM-based framework for knowledge transfer in the injection molding industry. The proposed system integrates both documented knowledge (as found in troubleshooting tables and manufacturing manuals) and extensive field data modeled using a diffusion generative model. This dual approach goes beyond previously trained general knowledge, enabling the delivery of precise, quantitative, and field-relevant expertise to support decision making in real industrial settings. Comparative validation using three representative OpenAI models demonstrated that more capable, recently developed models, particularly GPT-4o, produce more accurate and cost-effective responses in complex task scenarios. These results highlight the practical value of leveraging high-performance LLMs for intelligent industrial assistance.

In addition, to address the challenges of handling multilingual, ambiguous, or incomplete inputs in real industrial environments, our framework incorporates a dedicated input formatting stage. Specifically, a Translator module was employed to normalize queries into English, while the Task Formatter module leveraged prior conversational history to preserve context and generate well-structured tasks. In particular, input formatting was essential for integrating deep learning models such as the diffusion model, which require complete and well-defined input conditions. To this end, we implemented an interactive approach that guides users in providing the necessary input parameters during the dialogue. As a result, the proposed framework enables robust knowledge transfer across multiple languages while ensuring that users supply clear and complete inputs for tool usage.

For validation, we constructed a customized benchmark consisting of 160 question–answer pairs, specifically designed to evaluate both single tool-usage (100) and multi-tool-usage (60) scenarios. Comparative validation using three representative OpenAI models demonstrated that more capable, recently developed models, particularly GPT-4o, consistently achieved the highest accuracy, but this came at the expense of higher latency and computational cost. In contrast, GPT-4o-mini and GPT-3.5-turbo offered faster and more affordable responses, but with noticeable accuracy degradation. These results underscore the importance of selecting LLMs according to task requirements in practical application considering industrial settings. This flexibility in model selection is a potential benefit of adopting a multi-agentic approach, which is not easily achievable through fine-tuning single-model approaches.

Additionally, this study assesses the evaluation capability of the LLM-as-judge method for injection molding-related queries. The weak correlation between LLM-based and human expert evaluations indicate that LLM assessment often overemphasize fluency while failing to validate numerical reasoning, making it insufficient for complex, domain-sensitive tasks. Accordingly, expert involvement remains indispensable for ensuring reliable evaluation and supporting the practical application of this framework in technical manufacturing domains.

Compared with the fine-tuned single-agent baseline, the fine-tuned model was more efficient in terms of latency and pricing, as it could generate responses through a single invocation. However, its initial construction required additional time and computational resources, and its overall accuracy tended to degrade. In contrast, the multi-agent IM-Chat, which dynamically invoked trained generative models such as the diffusion model, demonstrated significantly higher accuracy in quantitative decision-making. Furthermore, when addressing multi-tool tasks that required the integration of information from heterogeneous sources, the multi-agent architecture of IM-Chat showed greater robustness and scalability, underscoring its potential advantages in complex, tool-integrated manufacturing environments.

Hence, beyond its immediate application, the IM-Chat framework offers strong potential for extensibility. By adopting a RAG approach instead of fine-tuning, the system remains modular and adaptable. New tools and pre-trained AI models can be easily integrated into the multi-agent architecture without modifying the core LLMs. While this work focused on the use of a diffusion model for quantitative inference, other generative or predictive models can be similarly incorporated, depending on the target domain. The flexible model integration capability of the framework enables its generalization beyond injection molding, and this study represents the first demonstration of such a system specifically designed for knowledge transfer in this domain.

However, several limitations remain. These can be categorized into three main areas: technical constraints, scalability challenges, and deployment considerations, which are discussed in detail below. First, technically, the current framework primarily operates on textual and structured data and has limited capabilities for extracting information from visual content such as images, tables, diagrams, and engineering drawings. These elements are frequently encountered in manufacturing documentation. While initial efforts were made to incorporate vision language models (VLMs) for parsing such content, the results revealed limitations in accuracy and completeness. In addition, even purely text-based documents often contain nonstandard formats, special symbols, or embedded figures that further hinder reliable parsing. These observations underscore the need for more robust multimodal capabilities that can jointly process both textual and visual information with higher precision.

In addition, the current system supports only a limited set of external tools. These primarily include troubleshooting tables and diffusion-based generative models. In real-world manufacturing environments, however, a much broader ecosystem of tools is used. These include empirical equation solvers, process simulators, optimization modules, and CAD-integrated analysis software. Such tools span the full lifecycle of industrial operations, from product design to failure analysis. To fully support these workflows, the IM-Chat framework must technically evolve into a generalized tool-calling infrastructure that is capable of interfacing with heterogeneous tools through modular adapters, dynamic invocation protocols, and error-tolerant input-output handling. This would allow intelligent agents to autonomously leverage diverse computational resources, thereby enabling deeper task specialization and more precise decision support across domains.

Second, with respect to scalability, the greatest hurdle for this framework lies in latency and pricing when ensuring accuracy, particularly when relying solely on multiple invocations of high-performance commercial

LLMs (e.g., GPT-4o). To address these constraints, several mitigation strategies can be considered. To begin with, selective routing of queries to lighter models can be applied for routine tasks, reserving high-capacity models for complex reasoning only. In addition, model compression and knowledge distillation can be employed to fine-tune small language models (SLMs) for specific tasks within the agentic workflow, which may incur higher initial development costs but yield long-term benefits. Finally, optimization of the multi-agent system can be pursued. Within the proposed architecture, three components in particular may be optimized: (i) the prompt design for each agent [122–124], (ii) the hyperparameters of the language models [125–127], and (iii) the coordination among agents [128–131]. The framework could further benefit from reinforcement learning and reward functions that explicitly account for latency and cost, guiding the system toward more efficient reasoning and more effective tool invocation.

Third, for deployment in the manufacturing field, several issues need to be considered. For instance, the quality of the system's response was found to be influenced by the specificity of user input. When user queries are vague or under-defined, even though we have integrated an input formatting stage, including the task formatting module and diffusion model–specific preprocessing, the system still tends to generate lower-quality outputs or, in some cases, fail to provide an answer. To mitigate this issue, future versions of the framework should incorporate more sophisticated input-to-query formatting modules. The formatting module could introduce an agent capable of iteratively checking whether key terms contain typographical errors or are contextually appropriate. In addition, if critical variables or essential elements are missing, the system could infer them by itself and propose candidate values to the user for confirmation, enabling a more interactive refinement process. This component would help users articulate their intent more clearly and improve the quality, relevance, and interpretability of downstream responses. Moreover, the use of external APIs for inference on proprietary process data raises concerns regarding intellectual property protection and data privacy, and may lead to vendor lock-in. Incorporating locally hosted SLMs for specific subtasks could not only improve cost efficiency and reduce latency, as mentioned above, but also enhance data privacy and mitigate vendor lock-in.

In summary, addressing current limitations such as restricted multimodal capabilities, scalability challenges related to latency and cost when relying on commercial LLMs, limited interoperability with external tools, and sensitivity to vague or underdefined user inputs will be essential for enhancing the robustness and practical value of the framework. As these aspects are progressively improved, the system has the potential to evolve into a scalable and general purpose solution for knowledge transfer in injection molding and related manufacturing domains.

## 5 CONCLUSION

This study presents IM-Chat, a multi-agent LLM framework that bridges documented knowledge sources and data-driven insights to support industrial knowledge transfer in the injection molding domain. The framework combines RAG with external tool integration, enabling it to process structured documents, utilize generative models, and deliver context-aware, task-specific responses.

Through extensive benchmarking, we demonstrated that high-performance models such as GPT-4o consistently outperform lighter alternatives in complex scenarios involving tool selection and numerical reasoning. While GPT-4o-mini and GPT-3.5-turbo offer advantages in terms of speed and cost, they often struggle with domain-sensitive subtasks that require deep reasoning or multi-step planning. This benchmarking thus offers practitioners actionable guidance for selecting and deploying LLM solutions in manufacturing contexts. In addition, we have shown that discrepancies between human expert and LLM-based evaluations, marked by low correlation scores, confirm that practitioner involvement is essential for reliable evaluation in technical manufacturing contexts.

In comparison with the fine-tuned single LLM baseline, IM-Chat achieves higher accuracy, especially in quantitative decision making based on the diffusion model, although it requires a longer reasoning time and higher computational cost. Furthremore, a key strength of IM-Chat lies in its modular and extensible architecture. This architecture enables seamless extension to additional models, effective integration of information from heterogeneous sources, and straightforward adaptation to different industrial domains with minimal adjustments. As a result, IM-Chat serves not only as a solution tailored to injection molding but also as a prototype for domain-specific AI systems applicable across the broader manufacturing sector.

Nonetheless, several limitations remain. While the system incorporates vision–language models to parse visual content such as diagrams, tables, and engineering drawings, its ability to extract and utilize such information remains limited and requires further improvement. The framework also depends heavily on commercial LLM APIs, which raises concerns regarding latency, cost, data privacy, and long-term sustainability. In addition, tool interoperability is still constrained, and the system's performance remains sensitive to the specificity and clarity of user queries.

To address these challenges, future research may explore the development of more robust multimodal capabilities, the expansion of tool interfaces to include simulators and engineering design software, and the use of input-refinement strategies to improve the clarity and relevance of user queries. In addition, lightweight task-specialized agents with selective fine-tuning could be investigated to reduce deployment overhead, while reinforcement learning strategies that explicitly account for latency and cost may offer promising directions for improving scalability and efficiency without sacrificing domain adaptability.

Collectively, this work illustrates the technical feasibility of applying multi-agent LLM systems in industrial contexts, while also highlighting remaining challenges related to technical constraints, scalability, and deployment. These findings suggest potential pathways toward future intelligent systems capable of adaptive reasoning, cross-domain generalization, and human-aligned interaction in manufacturing environments.

Supplementary Material

# IM-Chat: A Multi-agent LLM Framework Integrating Tool-Calling and Diffusion Modeling for Knowledge Transfer in Injection Molding Industry

Junhyeong Lee[1,†], Joon-Young Kim[1, 2,†], Heekyu Kim[1,†], Inhyo Lee[1] and Seunghwa Ryu[1,*]

**Affiliations**

[1]Department of Mechanical Engineering, Korea Advanced Institute of Science and Technology (KAIST), 291 Daehak-ro, Yuseong-gu, Daejeon 34141, Republic of Korea

[2]Industrial Intelligence Research Group, AI/DX Center, Institute for Advanced Engineering (IAE), Yongin, Republic of Korea

[*]Corresponding authors' e-mail: ryush@kaist.ac.kr

**Supplementary Note 1. IM-Chat Implementation and Prompt Engineering Details**

This section provides the complete prompt templates used to configure the core modules of the IM-Chat framework. To ensure transparency and reproducibility, each prompt is presented in its entirety, with variables dynamically inserted at runtime denoted by curly brackets (e.g., {input}). Accompanying explanations of these variables are included to enhance reader comprehension and enable accurate interpretation of the prompt structure.

In addition, for each module, we explicitly specify its role, task objective, operational constraints, and expected output format. This structured specification enables precise control over modules behavior and ensures consistent performance across varying task contexts. Notably, the design of each module is prompt-driven, allowing for flexible modification and rapid extension of system functionalities. This modularity underscores the adaptability and scalability of the IM-Chat framework in diverse manufacturing applications.

### 1.1 Prompt for *Task Formatter*

The *Task formatter* module is responsible for transforming the user's current input query into a structured format that clearly reflects the user's intent. To do so, it retrieves and incorporates relevant context from the accumulated **conversation_history**, enabling the system to disambiguate user instructions and infer implicit information. The output is a structured query that synthesizes both the new input and prior conversational knowledge, providing a precise and context-aware representation of the user's task request. This structured query serves as the foundation for subsequent tool selection and execution. The prompt used for the *Task formatter* is shown below.

```
## ROLE ##
```

You are a context-aware AI assistant specializing in extracting user intent and relevant information from conversation history.

---

## TASK ##

1. **Analyze** the conversation history and determine the user’s current request.

2. **Retrieve** all necessary details from the history that are relevant to answering the request.

3. **Format** the extracted information in a structured way to be used by a subsequent AI agent.

---

## ENSURE THAT ##

- The user’s intent is captured **clearly and concisely**.

- No important details from the conversation history are **missed**.

- The output is structured for easy parsing by another AI agent.

---

## OUTPUT FORMAT ##

- User's Current Request:

(Summarized user request)

- Relevant Information from Conversation History:

- (Relevant detail 1)

- (Relevant detail 2)

- (Relevant detail 3)

...

---

Then, Here is the conversation history (Please use the language the user is speaking):

```
{conversation_history}
```

### 1.2 Prompt for *Translator*

To accommodate the multilingual environments commonly encountered in the injection molding industry, where workers may use different native languages, a *Translator* module was introduced as part of the input formatting stage. This module automatically translates all user inputs into English, ensuring compatibility with downstream reasoning and tool modules that operates in a single language. In parallel, it identifies and records the user's original language, allowing the system to deliver responses back in the appropriate language when needed. This design enables seamless multilingual interaction while maintaining a consistent internal representation of user queries. The following prompt is used to confiture the *Translator* module.

```
You are a translation assistant.

Your task is to:

1. Detect the language of the input text.

2. If the input is already in English, do not translate it. Just return it as is.

3. If the input is not in English, translate it into English.

4. Return the result in the following format:

{format_instructions}

---

Input text:

{input}
```

The **format_instructions** schema employed in the *Translator* module is defined by the following class. It ensures that the translated query is returned in English, accompanied by metadata indicating the original input language.

```
class Translated(BaseModel):
    translated_query: str = Field(description="The translated input query in English.")
    language: str = Field(description="The original language of the input query.")
```

The **input** to the *Translator* module consists of a structured query, generated by the *Task Formatter*, in the language originally used by the user.

### 1.3 Prompt for *Classifier*

The *Classifier* module is designed to perform conditional routing based on the semantic category of the incoming user query. Specifically, it determines whether the query is injection molding-related or not. Queries classified as injection-related are processed through a structured plan-and-execute pipeline, which carefully invokes appropriate tools to generate accurate responses. In contrast, queries deemed unrelated to injection molding are delegated to a lightweight ReAct-style agent that leverages web search without invoking specialized tools. Additionally, queries that can be resolved solely through retrieval from the existing conversation history are also categorized as non-injection-related. The output of the classifier is a binary label: either "*injection*" or "*no_injection*". The corresponding system prompt used to implement this classification logic is provided below.

```
## ROLE ##
```

You classify **injection molding queries** based on whether they require tool-based processing.

---

## CLASSIFICATION RULES ##

- **`injection`** → If the query is:

  1. **Directly related to injection molding** (e.g., defects, machine maintenance, material settings).

  2. **Requires at least one available tool** to generate a solution.

- **`no_injection`** → If the query:

  - Is unrelated to injection molding.

  - Covers general knowledge, theory, or industry trends.

  - Can be answered **fully using conversation history** (without tools).

---

## TOOLS ##

If classified as **`injection`**, process using:

- **table_retriever** → Retrieves **senior worker-documented process parameter adjustments** for each defect type from a predefined table.

- **manual_retriever** → Retrieves  information about inejction molding machine: Machine Maintenance, Machine Inspection, Machine Operation, and Material settings etc.

- **internet_search** → Finds **external information** only if relevant information is not available in the manual or table.

- **llm_infer** → Relies solely on internal reasoning without using any external tools..

- **diffusion_model** → Exclusively used to generate injection molding process parameter sets using a diffusion model, based on environmental variables provided by the user. Diffusion model needs 4 parameters: Machine temperature, Machine humidity, Factory temperature, and Factory humidity.

**Do not** provide explanations. If uncertain, default to `no_injection`.

In the *Classifier* module, the parsing schema is implemented using the BaseModel class and incldes the use of Literal types to constrain specific field values, as detailed below.

### 1.4 Prompt for *ReAct* Agent (agent for non-injection molding-related tasks)

The *ReAct* agent represents a representative class of reasoning agents that iteratively alternate between thought generation and tool invocation to solve complex tasks. In the IM-Chat framework, it is employed to efficiently handle queries unrelated to injection molding, allowing for lightweight and rapid response generation without engaging the full plan-and-execute pipeline. The system prompt used to configure the *ReAct* agent is provided below.

## ROLE ##

You provide **clear, accurate, and user-friendly answers** on general topics.

Adapt your tone to match the user's style when appropriate.

## TOOL USAGE ##

Use **"internet_search"** **ONLY IF**:

- The information is **time-sensitive** (e.g., news, weather, recent tech updates).

- The answer is **outside your knowledge scope**.

- External confirmation is **necessary for accuracy**.

If using **"internet_search"**, summarize findings and cite key sources.

## OUTPUT EXPECTATIONS ##

- Provide **complete** answers to minimize follow-up questions.

- Use **step-by-step explanations** when helpful.

- Break complex topics into sections for clarity.

- Ensure **accuracy, clarity, and helpfulness** in all responses.

---

Answer the question as thoroughly as possible, using tools only when necessary.

### 1.5 Prompt for *Planner*

The *Planner* module is responsible for decomposing a given query related to injection molding into a sequence of actionable subtasks to facilitate solution generation. Upon receiving the input query ({input}), the *Planner* considers the scope of executable actions available to the *Executor* and outputs a structured plan following the schema specified in {format_instructions}.The full system prompt used for the *Planner* is provided below.

## ROLE ##

You are the **Planner, responsible for creating **minimal, step-by-step advisory plans** for **injection molding queries**.

Each step must be **essential, self-contained, and lead efficiently to the final conclusion**.

---

## GUIDELINES ##

- **Use the fewest steps possible to reach a conclusion.**

- **Each step must provide direct, actionable guidance.**

- **Avoid unnecessary details, extra interpretation, or redundant steps.**

- **Ensure the final step gives a clear answer.**

---

## AVAILABLE TOOLS ##

- **table_retriever** → Retrieves **senior worker-documented process parameter adjustments** for each defect type from a predefined table.

- **manual_retriever** → Retrieves information about inejction molding machine: Machine Maintenance, Machine Inspection, Machine Operation, and Material settings etc.

- **internet_search** → Finds **external information** only if relevant information is not available in the manual or table.

- **llm_infer** → Relies solely on internal reasoning without using any external tools..

- **diffusion_model** → Exclusively used to generate injection molding process parameter sets using a diffusion model, based on environmental variables provided by the user. Diffusion model needs 4 parameters: Machine temperature, Machine humidity, Factory temperature, and Factory humidity.

**Note:**

- The information obtained from tools is **accurate and does not require review or further analysis steps**. Simply use the retrieved data as is.

```
**Focus on the fastest path to a useful advice—no unnecessary steps.**

---

## INPUTS ##

- Input query: {input}

---

## OUTPUT FORMAT ##

{format_instructions}
```

The planning schema is defined by the *Plan* class, implemented using BaseModel. It includes a single field, *steps*, which is a list of string representing the ordered sequence of tool-invocation steps. Each element in the list corresponds to a tuple specifying the tool to be used and a brief description of the associated subtasks.

```
class Plan(BaseModel):
    steps: List[str] = Field(
        description="""
        A list of ordered steps to follow. Each step is a tuple in the format:
        [(Tool to be used, Task description),
         (Tool to be used, Task description),
         ...]

        Steps must be ordered sequentially.
        """
    )
```

## 1.6 Prompt for *Executor*

The *Executor* module is designed to operate on the first subtask provided by the *Plan*. Only this initial subtask is passed as input, and the *Executor* dynamically selects the appropriate tool to accomplish the task. To handle execution, we employed a *ReAct*-style agent capable of invoking up to five tools, including LLM-based inference. The corresponding prompt used to configure this module is presented below.

## ROLE ##

You are the **Executor Agent**, responsible for executing **injection molding tasks** as instructed by the Planner.

Execute the task given to you with using the proper tool (mentioned on the task).

---

## AVAILABLE TOOLS ##

- **table_retriever** → Retrieves **senior worker-documented process parameter adjustments** for each defect type from a predefined table.

- **manual_retriever** → Retrieves  information about inejction molding machine: Machine Maintenance, Machine Inspection, Machine Operation, and Materials Settings etc.

- **internet_search** → Finds **external information** only if relevant information is not available in the manual or table.

- **llm_infer** → Relies solely on internal reasoning without using any external tools..

- **diffusion_model** → Exclusively used to generate injection molding process parameter sets using a diffusion model, based on environmental variables provided by the user. Diffusion model needs 4 parameters: Machine temperature, Machine humidity, Factory temperature, and Factory humidity.

---

## IMPORTANT ##

- **Do not anticipate future steps.**

- **Do not skip or combine steps.**

- **Use the tool ONLY ONCE and give the resopnse.**

- **If the task is not clearly executable or falls outside your capabilities, respond exactly with:**

  `"This is out of my scope. Please specify the task."`

### 1.7 Prompt for *Supervisor*

The *Supervisor* module is responsible for determining whether the current solution provided by the *Executor*, along with the accumulated solution history, is sufficient to fully resolve the original input query. If the available information is deemed adequate, the *Supervisor* returns the signal "respond" to indicate that a final answer can be generated. Conversely, if the solution is incomplete or insufficient, the Supervisor outputs "replan" to trigger another round of planning and execution. To support this decision-making process, the Supervisor prompt is supplied with the original query ({input}), the current plan ({plan}), and the execution history ({past_steps}), which includes all prior subtasks and their corresponding results. The complete prompt is shown below.

You are a decision-making agent.

Your task is to read the following input, and based on the conditions, retrun either "replan" or "respond" as your decision.

---

## INPUTS ##

- Input query: {input}

- Original plan: {plan}

- Completed Steps: {past_steps}

---

## OPERATION MODE ##

Determine your role based on the following rules:

1. If the **Original Plan** is incomplete or unclear, or if **Completed Steps** do not yet fulfill the Objective**, then:

→ Output: "replan"

2. If the **Completed Steps are sufficient to achieve the Objective**, then:

→ Output: "respond"

3. If **necessary information is missing and the Objective cannot be completed without it**, then: Especially for diffusion model, we strictly need four inputs: Machine temperature, Machine humidity, Factory temperature, and Factory humidity.

→ Output: "respond"

**Important Notes:**

- Do NOT try to guess or assume missing information.

- Base your decision solely on "INPUTS" provided.

## OUTPUT FORMAT ##

{format_instructions}

The format_instructions schema is implemented using a BaseModel with a Literal type to ensure that the output of the *Supervisor* is restricted to either "respond" or "replan" as valid decisions, as shown below.

```
class Supervisor_decision(BaseModel):
    decision: Literal["replan", "respond"]
```

### 1.8 Prompt for *Replanner*

The *Replanner* module functions similarly to the original *Planner*, generating an actionable plan for the *Executor* to follow. However, unlike the *Planner*, it revises the plan by incorporating not only the initial input query but also the existing plan and the execution history provided by the *Executor*. To enable this, the prompt includes {input}, {plan}, and {past_steps}, as in the *Supervisor* module. The output is structured using the same *Plan* class defined in the *Planner* module. The prompt is provided below.

```
## ROLE ##

You are the **Replanner Agent**, responsible for updating a step-by-step plan or finalizing the
planning process and responding to the user.

You are part of a knowledge transfer system.

Your role is to deliver accurate, structured knowledge related to the given objective — not to
modify the objective or take control of the process yourself.

You must decide whether to:

- Revise and continue the plan, if it is incomplete or suboptimal.
```

- Finalize and respond, if all necessary steps have been completed.

## GUIDELINES ##

- **Use the fewest steps possible to reach a clear and useful conclusion.**

- **Avoid unnecessary details, redundant logic, or excessive interpretation.**

- **The final step must clearly address the objective.**

## AVAILABLE TOOLS ##

- **table_retriever** → Retrieves **senior worker-documented process parameter adjustments** for each defect type from a predefined table.

- **manual_retriever** → Retrieves information about inejction molding machine: Machine Maintenance, Machine Inspection, Machine Operation, and Material settings etc.

- **internet_search** → Finds **external information** only if relevant information is not available in the manual or table.

- **llm_infer** → Relies solely on internal reasoning without using any external tools..

- **diffusion_model** → Exclusively used to generate injection molding process parameter sets using a diffusion model, based on environmental variables provided by the user. Diffusion model needs 4 parameters: Machine temperature, Machine humidity, Factory temperature, and Factory humidity.

**Note:**

- Information retrieved via tools is **accurate** and should be used **as is** — no need for validation or additional analysis.

**Your goal is to provide the fastest possible path to useful, actionable knowledge.**

---

## INPUTS ##

- Input query: {input}

- Original plan: {plan}

- Completed Steps: {past_steps}

---

## OUTPUT FORMAT ##

{format_instructions}

### 1.9 Prompt for *Reporter* (for non-injection tasks)

The *Reporter* module is designed to reformat and summarize the solution, originally generated in English, into the user's native language for final display. Given the English query {input_eng} and its corresponding response {response}, the module utilizes the detected user language {language}, as identified by the *Translator* module, to generate a localized and user-friendly summary. This ensures that the final output is both linguistically accessible and contextually coherent for multilingual users. The full prompt is provided below.

You are the final output formatting agent in a multi-agent system. Your role is to present the final response in a clear and natural format suitable for a {language}-speaking user.

The original user query is:

{input_eng}

The final response for this query is:

{response}

Please follow these guidelines:

1. Regardless of the original input language, make sure the final response is clearly understandable in {language}, preserving the original meaning, tone, and nuance.
2. Do not skip or delete content, even if it seems already understandable. Your job is to make sure the user receives a complete, well-formatted, and respectful response in their language.
3. You may retain certain English words if doing so improves clarity or intent, consdiering user query and the response.
4. Do not summarize, omit, or rephrase content unless it's truly necessary for natural readability in {language}.

Present only the final, user-facing response, without explanation or markup.

Remember: your job is not just translation — it's to deliver the final output as if a human assistant were speaking directly to a {language}-speaking user.

### 1.10 Prompt for *Reporter* (for injection-related tasks)

For injection-related tasks, the *Reporter* module is configured to generate a concise and user-friendly summary based not only on the final solution, but also on the full execution history captured in {past_steps}. This input includes the sequence of subtasks executed throughout the plan-and-execute loop, along with the associated tool outputs. The prompt is designed to distill this information into a coherent response suitable for end-user communication. The complete prompt is presented below.

```
The original user query is:
{input_eng}

To address this query, we referred to the following reasoning steps:
{past_steps}

Please do the following:

1. Based on the reasoning steps, extract and summarize only the most relevant information
necessary to answer the query.
2. Translate the explanation into {language}, preserving the meaning and tone. Presnet only the
translated explanation.
- If the language is English, skip the translation step silently.
- Use the original English term in parentheses when translating technical words.

Keep the final output polite, context-aware, and easy to understand. Avoid unnecessary
elaboration.
```

**Supplementary Note 2. Prompt Details for Retrieval Methods**

For tasks related to injection molding, the Executor leverages a set of retriever tools to augment responses with factual information drawn from domain-specific knowledge sources. These tools retrieve semantically relevant chunks by performing embedding-based similarity search against the input query. Rather than simply returning the matched chunks, the retrieved information is post-processed—summarized and reformatted—into a structured form that facilitates integration into the reasoning workflow. This approach ensures that the Executor receives well-organized and contextually relevant knowledge to support informed decision-making. This section presents the full prompts used for the retrievers responsible for accessing and synthesizing information from both the troubleshooting table and the manufacturing manual.

**2.1 Prompt for Troubleshooting Table Retrieval Tool**

## ROLE ##

Given a specific defect type, extract and organize the corresponding process parameters and their adjustment directions & orders **strictly based on the information provided in the DOCUMENT section**.

- Use ONLY the content from DOCUMENT.

- Do NOT make assumptions or add any suggestions.

---

## DOCUMENT ##

You will receive two documents:

1. A table indicating whether each process parameter should be increased ("+") or decreased ("-") to resolve the given defect.

2. A table specifying the adjustment priority (lower numbers indicate higher priority).

Documents:

{context}

---

## OUTPUT FORMAT ##

### **Parameter Adjustments (sorted by adjustment order):**

**(Priority: 1)** → (Parameter, Increase / Decrease), (Parameter, Increase / Decrease), ...

**(Priority: 2)** → (Parameter, Increase / Decrease), (Parameter, Increase / Decrease), ...

... (continue as needed) ...

---

### **If the defect type is ambiguous or unspecified:**

Respond with:

"I cannot answer that query because the defect type is not specific."

**2.2 Prompt for Machine Manual Retrieval Tool**

## ROLE ##

You are an information extractor about the query about **injection molding machine operation & maintenance**

Use ONLY the provided manual. **DO NOT assume or invent information.**

## GUIDELINES ##

- Answer strictly based on the manual.

- Indicate whether your answer is from by page number.

## DOCUMENT ##

{context}

## OUTPUT FORMAT ##

**Answer:** [Provide response based on the manual]

**Reference:** See [page number] for detail.

### **If you cannot find related information from the document:**

Respond with:

"The manual does not contain the information you mentioned about."

**Supplementary Note 3. Technical Description of the Classifier-Free Guidance Diffusion Model (CFGDM)**

The diffusion model aims to learn a reverse process that reconstructs the original data distribution from a noisy prior, enabling new sample generation. Initially, noise is incrementally introduced to the original data, converting it into a noisy prior (forward process). The model is then trained to progressively denoise this prior and recover the original distribution (reverse process). Both processes proceed through incremental steps defined by a Markovian chain, where each state depends solely on its preceding state. In this scenario, denoting each time step as t (where t=0,…,T) , both forward and reverse processes, defined by conditional distributions based on a Markovian chain, follow Gaussian distributions.

Formally, using Bayes' rule, the reverse process can be expressed as:

$$q(\mathbf{X}_{t-1}|\mathbf{X}_t) = \frac{q(\mathbf{X}_t|\mathbf{X}_{t-1})q(\mathbf{X}_{t-1})}{q(\mathbf{X}_t)} \tag{1}$$

Here, $q(\mathbf{X}_t)$ indicates the distribution of the sample $\mathbf{X}$ corresponding to the specific time step $t$. In **Eq. 1**, direct estimation of $q(\mathbf{X}_t)$ and $q(\mathbf{X}_{t-1})$ is infeasible, preventing computation of $q(\mathbf{X}_{t-1}|\mathbf{X}_t)$. To resolve this, an approximation function $p_\theta(\mathbf{X}_{t-1}|\mathbf{X}_t)$ is introduced and trained via machine learning methods. Consequently, the diffusion model is optimized to fulfill the subsequent condition.

$$p_\theta(\mathbf{X}_{t-1}|\mathbf{X}_t) \approx q(\mathbf{X}_{t-1}|\mathbf{X}_t) \tag{2}$$

### 3.1 Forward Process

In the forward process, Gaussian noise is incrementally added based on a predefined variance schedule $\beta_t$, typically using linear or cosine schedules. The conditional probability distribution is:

$$q(\mathbf{X}_t|\mathbf{X}_{t-1}) \coloneqq N\left(\mathbf{X}_t\,;\ \boldsymbol{\mu}_{\mathbf{X}_{t-1}}, \boldsymbol{\Sigma}_{\mathbf{X}_{t-1}}\right) \coloneqq N\left(\mathbf{X}_t\,; \sqrt{1-\beta_t}\mathbf{X}_{t-1},\ \beta_t\mathbf{I}\right) \tag{3}$$

$\beta_t$ governs the level of noise added and is commonly adjusted according to the time step $t$ using various scheduling techniques. For gradient computation, noise addition is reformulated using the reparameterization trick, similar to the methods used in variational autoencoder (VAE) [1], resulting in a closed-form expression:

$$q(\mathbf{X}_t|\mathbf{X}_{t-1}) = \sqrt{1-\beta_t}\mathbf{X}_{t-1} + \sqrt{\beta_t}\boldsymbol{\epsilon}_{t-1}, \qquad \text{where } \boldsymbol{\epsilon} \sim N(\mathbf{0},\ \mathbf{I}) \tag{4}$$

Using the Markov property, the forward process up to any chosen step $T$ can be represented as the sequential multiplication of conditional probabilities, formulated as:

$$q(\mathbf{X}_{1:T}|\mathbf{X}_0) \coloneqq \prod_{t=1}^{T} q(\mathbf{X}_t|\mathbf{X}_{t-1}) \tag{5}$$

Moreover, by assuming that the variables $\boldsymbol{\epsilon}_{t-1}, \boldsymbol{\epsilon}_{t-2}, \ldots, \boldsymbol{\epsilon}_0$ follow an independent Gaussian distribution $N(\mathbf{0},\ \mathbf{I})$, and utilizing the property that the sum of two Gaussian distributions $N(\mathbf{0},\ \sigma_1^2\mathbf{I})$ and $N(\mathbf{0},\ \sigma_2^2\mathbf{I})$ yields another Gaussian distribution $N(\mathbf{0},\ (\sigma_1^2+\sigma_2^2)\mathbf{I})$, the sampling at any given timestep $t$ can be expressed analytically:

$$\mathbf{X}_t = \sqrt{\overline{\alpha}_t}\mathbf{X}_0 + \sqrt{1-\overline{\alpha}_t}\boldsymbol{\epsilon} \tag{6}$$

Here, $\alpha_t$ is defined as $1-\beta_t$, and $\overline{\alpha}_t$ as $\prod_{i=1}^{t}\alpha_i$, while $\overline{\boldsymbol{\epsilon}}$ denotes the standard deviation of the merged Gaussian distribution.

### 3.2 Reverse Process

To train the diffusion model, the goal is to maximize the negative log-likelihood of the original data $\mathbf{X_0}$. However, since $\boldsymbol{p_\theta}(\mathbf{X_0})$ cannot be directly evaluated, the model instead leverages a variational lower bound (VLB), a strategy also used in VAE:

$$\begin{aligned} -\log p_\theta(\mathbf{X}_0) &\leq -\log p_\theta(\mathbf{X}_0) + D_{KL}\big(q(\mathbf{X}_{1:T}|\boldsymbol{X}_0)|p_\theta(\mathbf{X}_{1:T}|\mathbf{X}_0)\big) \\ &= \mathbb{E}\left[\log \frac{q(\mathbf{X}_{1:T}|\mathbf{X}_0)}{p_\theta(\mathbf{X}_{0:T})}\right] \end{aligned} \tag{7}$$

$$= \mathbf{L}_{\text{VLB}}$$

Sohl-Dickstein et al. [2] made the above objective more analytically tractable by decomposing it into three components:

$$\mathbf{L}_{\text{VLB}} = \mathbb{E}_q \left[ D_{KL}\big(q(\mathbf{X}_T|\mathbf{X}_0) \parallel p_\theta(\mathbf{X}_T)\big) + \sum_{t=2}^{T} D_{KL}\big(q(\mathbf{X}_{t-1}|\mathbf{X}_t, \mathbf{X}_0) \parallel p_\theta(\mathbf{X}_{t-1}|\mathbf{X}_t)\big) - \log p_\theta(\mathbf{X}_0|\mathbf{X}_1) \right] \tag{8}$$

Let, $\mathbf{L}_T \coloneqq D_{KL}(q(\mathbf{X}_T|\mathbf{X}_0) \parallel p_\theta(\mathbf{X}_T))$, $\mathbf{L}_t \coloneqq D_{KL}\big(q(\mathbf{X}_{t-1}|\mathbf{X}_t, \mathbf{X}_0) \parallel p_\theta(\mathbf{X}_{t-1}|\mathbf{X}_t)\big)$ $(1 \le t \le T-1)$, and $\mathbf{L}_0 \coloneqq -\log p_\theta(\mathbf{X}_0|\mathbf{X}_1)$. Then, the variational lower bound $\mathbf{L}_{\text{VLB}}$ becomes:

$$\mathbf{L}_{\text{VLB}} = \mathbf{L}_T + \mathbf{L}_{T-1} + \cdots + \mathbf{L}_0 \tag{9}$$

Among these terms, $\mathbf{L}_T$ is associated with the noise schedule $\beta_t$ in the forward process, which is predefined—typically via linear or cosine scheduling—and not subject to learning. Additionally, since $p_\theta(\mathbf{X}_T)$ represents Gaussian noise and contains no learnable parameters, $\mathbf{L}_T$ remains fixed during training and can be ignored. Similarly, $\mathbf{L}_0$ has been shown to have negligible impact on model performance and is typically omitted from the training objective [3].

The core objective thus reduces to optimizing $\mathbf{L}_t$ for $1 \le t \le T-1$. When $\beta_t$ is small, the forward and reverse processes can be approximated by similar Gaussian conditionals [2], allowing:

$$p_\theta(\mathbf{X}_{t-1}|\mathbf{X}_t) = N\big(\mathbf{X}_{t-1};\ \boldsymbol{\mu}_\theta(\mathbf{X}_t, t),\ \boldsymbol{\Sigma}_\theta(\mathbf{X}_t, t)\big) \text{ for } 1 < t \le T \tag{10}$$

Here, $\boldsymbol{\Sigma}_\theta(\mathbf{X}_t, t) = \sigma_t^2\mathbf{I}$ is set as untrained time dependent constants. Though $q(\mathbf{X}_{t-1}|\mathbf{X}_t)$ is intractable, its conditional form given $\mathbf{X}_0$ is analytically accessible:

$$q(\mathbf{X}_{t-1}|\mathbf{X}_t, \mathbf{X}_0) = N(\mathbf{X}_{t-1}; \boldsymbol{\mu}(\mathbf{X}_t,\ \mathbf{X}_0), \beta_t\mathbf{I}) \tag{11}$$

Using Bayes' rule and properties of Gaussian distributions, the mean $\tilde{\boldsymbol{\mu}}_t$ can be derived as:

$$\tilde{\boldsymbol{\mu}}_t(\mathbf{X}_t,\ \mathbf{X}_0) = \frac{\sqrt{\alpha_t}(1 - \bar{\alpha}_{t-1})}{1 - \bar{\alpha}_t}\mathbf{X}_t + \frac{\sqrt{\bar{\alpha}_{t-1}}\beta_t}{1 - \bar{\alpha}_t}\mathbf{X}_0 \tag{12}$$

From Eq. 6, we know:

$$\mathbf{X}_0 = \frac{1}{\sqrt{\bar{\alpha}_t}}\left(\mathbf{X}_t - \sqrt{1-\bar{\alpha}_t}\boldsymbol{\epsilon}_t\right) \tag{13}$$

Substituting this into the expression for $\tilde{\boldsymbol{\mu}}_t$, we get:

$$\tilde{\boldsymbol{\mu}}_t(\mathbf{X}_t, \mathbf{X}_0) = \frac{1}{\sqrt{\alpha_t}}\left(\mathbf{X}_t - \frac{1-\alpha_t}{\sqrt{1-\bar{\alpha}_t}}\boldsymbol{\epsilon}_t\right) \tag{14}$$

This leads to the following loss function:

$$\mathbf{L}_t = \mathbb{E}_q\left[\frac{1}{2\sigma_t^2}\|\tilde{\boldsymbol{\mu}}_t(\mathbf{X}_t, \mathbf{X}_0) - \boldsymbol{\mu}_\theta(\mathbf{X}_t, t)\|^2\right] + \mathbf{C} \tag{15}$$

Here, $\mathbf{C}$ is a constant term that does not depend on $\theta$. The function $\boldsymbol{\mu}_\theta(\mathbf{X}_t, t)$ serves as a natural candidate for estimating the target mean $\tilde{\boldsymbol{\mu}}_t(\mathbf{X}_t, \mathbf{X}_0)$. Given that $\boldsymbol{\mu}_\theta$ is trained to approximate $\tilde{\boldsymbol{\mu}}_t(\mathbf{X}_t, \mathbf{X}_0) = \frac{1}{\sqrt{\alpha_t}}\left(\mathbf{X}_t - \frac{1-\alpha_t}{\sqrt{1-\bar{\alpha}_t}}\boldsymbol{\epsilon}_t\right)$, it can be reparameterized accordingly:

$$\boldsymbol{\mu}_\theta(\mathbf{X}_t, t) = \frac{1}{\sqrt{\alpha_t}}\left(\mathbf{X}_t - \frac{\beta_t}{\sqrt{1-\bar{\alpha}_t}}\boldsymbol{\epsilon}_\theta(\mathbf{X}_t, t)\right) \tag{16}$$

$\boldsymbol{\epsilon}_\theta$ is a neural network that predicts the noise $\boldsymbol{\epsilon}$ given $\mathbf{X}_t$. By inserting **Eq. 14** and **Eq. 16** into **Eq. 15**, the loss expression can be reformulated as follows:

$$\mathbf{L}_t = \mathbb{E}_{\mathbf{X}_0, \boldsymbol{\epsilon}}\left[\frac{\beta_t^2}{2\sigma_t^2\alpha_t(1-\bar{\alpha}_t)}\left\|\boldsymbol{\epsilon} - \boldsymbol{\epsilon}_\theta\left(\sqrt{\bar{\alpha}_t}\mathbf{X}_0 + \sqrt{1-\bar{\alpha}_t}\boldsymbol{\epsilon}, t\right)\right\|^2\right] \tag{17}$$

Since constant factors do not affect optimization, the practical loss function simplifies to:

$$\left\|\boldsymbol{\epsilon} - \boldsymbol{\epsilon}_\theta\left(\sqrt{\bar{\alpha}_t}\mathbf{X}_0 + \sqrt{1-\bar{\alpha}_t}\boldsymbol{\epsilon}, t\right)\right\| \tag{18}$$

Once training is complete, new samples can be generated iteratively using:

$$\mathbf{X}_{t-1} = \frac{1}{\sqrt{\alpha_t}}\left(\mathbf{X}_t - \frac{\beta_t}{\sqrt{1-\bar{\alpha}_t}}\boldsymbol{\epsilon}_\theta(\mathbf{X}_t, t)\right) + \sigma_t\mathbf{z}, \quad \mathbf{z} \sim N(\mathbf{0}, \mathbf{I}) \tag{19}$$

### 3.3 Classifier-Free Guidance Diffusion Model (CFGDM)

The previously described diffusion model allows for the generation of process parameters that resemble those found in the training data. However, in an unconditional setting, the model simultaneously learns the distributions corresponding to both non defective and defective products, which can make it difficult to accurately capture the data distribution. Prior research [4,5] has shown that introducing class labels as conditioning information can improve generative model training by enabling better separation of different data categories. Moreover, in real-world scenarios, external factors such as factory temperature and humidity, as well as the temperature and humidity of materials, are uncontrollable. Therefore, it is essential to generate parameters for non defective products under given environmental conditions.

To address this, the Classifier-Free Guidance Diffusion Model (CFGDM) [6] is employed. Unlike the classifier guidance network introduced by Dhariwal et al. [7], CFGDM achieves conditional generation without requiring an external classifier, allowing for easier integration into existing diffusion frameworks. This approach jointly optimizes conditional and unconditional objectives by randomly dropping the condition **y** during training with a certain probability. The modified prediction is computed as:

$$\check{\boldsymbol{\epsilon}}(\mathbf{z}_t, \mathbf{y}, t) = w\boldsymbol{\epsilon}(\mathbf{z}_t, \mathbf{y}, t) + (1 - w)\boldsymbol{\epsilon}(\mathbf{z}_t, t) \tag{20}$$

Here, $w$ controls the influence of the conditional prediction, while $\boldsymbol{\epsilon}(\mathbf{z}_t, \mathbf{y}, t)$ and $\boldsymbol{\epsilon}(\mathbf{z}_t, t)$ refer to the outputs of the conditional and unconditional models, respectively. A study by Ho et al. [6] explored optimal values for the condition drop rate and guidance strength, recommending a drop probability of 0.1 and setting $w$ between 0.1 and 4. Following this guidance, we adopt a drop rate of 0.1 and use $w = 3$ in our implementation.

The training procedure can be outlined in the following steps:

1. Perturbed process parameters at time step $t$ are generated by injecting noise as defined in **Eq. 6.**

2. These perturbed parameters are fed into a U-Net, which outputs both the conditional noise prediction $\boldsymbol{\epsilon}(\mathbf{z}_t, \mathbf{y}, t)$ and the unconditional prediction $\boldsymbol{\epsilon}(\mathbf{z}_t, t)$. These are then combined to compute $\check{\boldsymbol{\epsilon}}(\mathbf{z}_t, \mathbf{y}, t)$ using **Eq. 20.**
3. The model is then optimized based on the loss function specified in **Eq. 18.**

### 3.4 Model Architecture and Training Setup

A U-Net architecture was employed, and the implementation provided in the PyTorch library [8] was adapted and modified for this study. The detailed architecture is illustrated in **Figure S1**.

Factory temperature, factory humidity, machine temperature, machine humidity, and product quality labels (good/defective) are used as conditioning variables. The process parameters to be generated include three injection speeds, three injection pressures, three injection positions, and one hold time, totaling 10 variables. These parameters, shaped as (batch size, 1, 10), are provided as inputs to the U-Net. The inputs first pass through a 1D convolution layer, resulting in a shape of (batch size, 64, 10). During downsampling, a structure consisting of two consecutive ResNet blocks and one attention block is repeated. Except for the first stage, the number of channels is doubled while the feature size is halved. In the final downsampling stage, only the number of channels is doubled, with the feature size preserved. The features then pass sequentially through a ResNet block, an attention block, and another ResNet block before entering the upsampling phase. In upsampling, the number of channels is halved and the feature size doubled, with skip connections linking the corresponding downsampling and upsampling features. After all upsampling steps, the features pass through a final 1D convolution layer, restoring the size to (batch size, 1, 10).

As shown in **Figure S1b**, conditioning information is injected through the ResNet blocks. The conditioning variables include quality labels (non-defective/defective) and four environmental factors (machine temperature, machine humidity, factory temperature, and factory humidity). The labels are first transformed through a class embedding layer, after which the transformed labels are concatenated with the environmental conditions and passed through a classes MLP. This output is subsequently merged with time information obtained via a time embedding layer. The combined information is processed by a condition MLP, decomposed into scale and shift components, and applied to the features via elementwise scaling and shifting (denoted as x in Figure S1b). A final 1D convolution layer produces the condition-encoded features.

The training configuration of the diffusion model is summarized in Table S1. Training was performed on NVIDIA RTX 3090 GPUs using the Adam optimizer with a learning rate of 1e-4. The model was trained for 100,000 steps, corresponding to approximately 2,380 epochs.

**Table S1:** The hyperparameters and training configuration of CFGDM used in this study.

| Diffusion model parameters | |
|---|---|
| Input size | 10 |
| Condition drop probability | 0.1 |
| Step size ($T$) | 1000 |
| Variance schedule (t) | Cosine |
| Optimizer | Adam; betas = (0.9, 0.99) |
| Learning rate | 1e-4 |
| Batch size | 64 |
| Training steps | 100,000 (2,380 epochs) |
| U-Net parameters | |
| Dimension | 64 |
| Dimension multipliers | 1, 2, 4, 8 |
| Attention heads | 4 |
| Dimension of attention head | 32 |
| ResNet block groups | 8 |

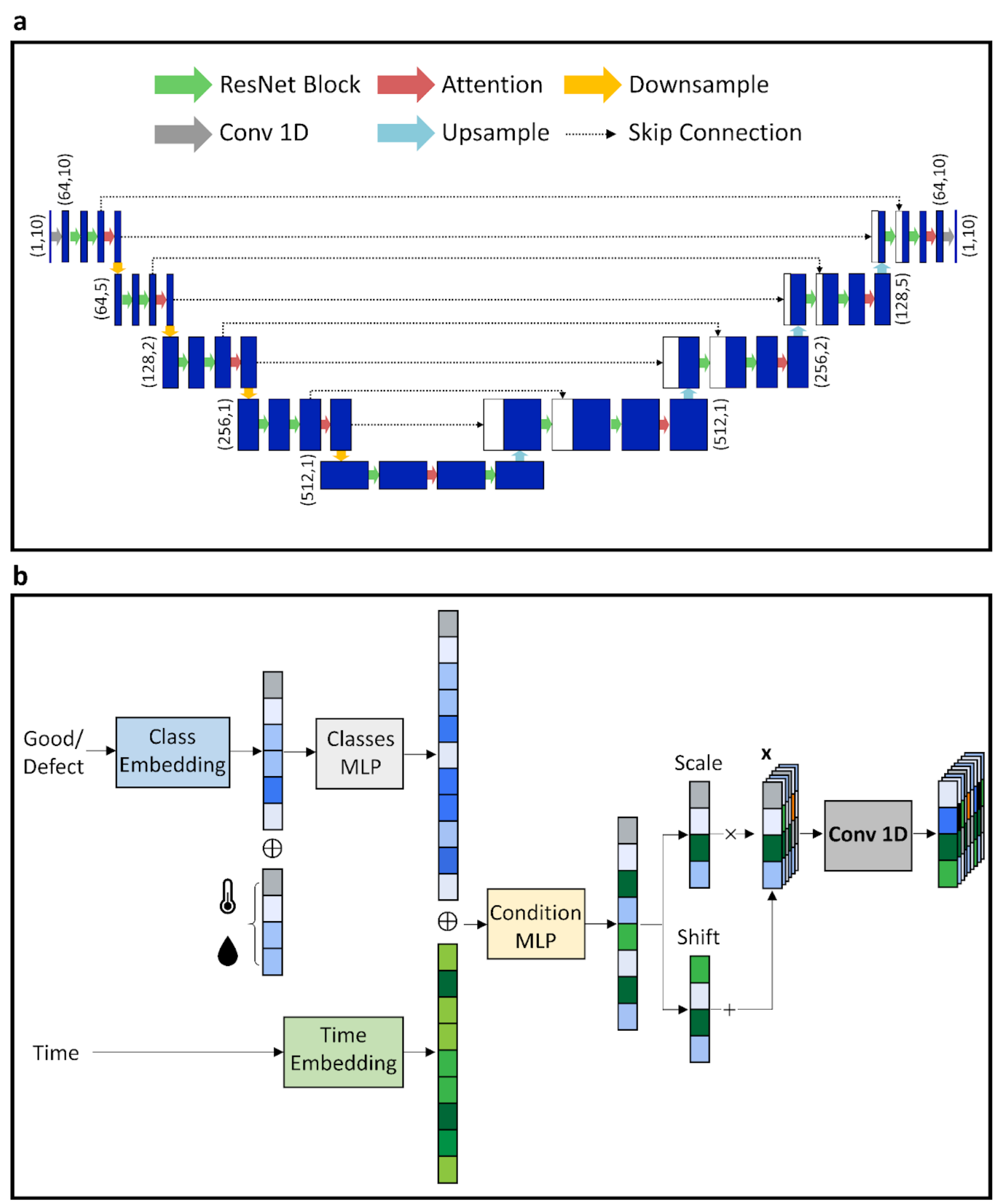

**Figure S1:** The model architecture of CFGDM

## Supplementary Note 4. Prompts for *Diffusion Input Formatter*

To interface with the diffusion model, five numerical input features are required: four environmental variables, machine temperature, machine humidity, factory temperature, and factory humidity, and a product quality class indicator (i.e., good or defective). Since the initial user queries are expressed in natural language, an input formatting module is necessary to extract and structure these variables appropriately. To achieve this, an LLM-based formatter is employed, guided by a prompt designed to infer the required parameters from the query. The system prompt is as follows:

```
You are a helpful assistant. For each input query, return a dictionary with exactly 5 keys:
'machine_temperature (Celcius degree)', 'machine_humidity (%)', 'factory_temperature (Celcius
degree)', 'factory_humidity (%)', and 'class (0 or 1)'.
Each key must have either a floating-point number as its value, or 'None' if there is not enough
information to determine a value.
You can keep the value as "None" if you do not have a clear and sufficient information to fill it in.
For the class, unless there are explicit incidents, let it be marked as good.
Please provide the output in this format for the input query.
```

The output of this module is parsed using a structured JSON schema, which ensures compatibility with the downstream diffusion model. The schema is defined as follows:

```
function_input_format = {
    "name": "formatter",
    "type": "function",
```

```
    "description": "Try to infer 5 properties: machine_temperature, machine_humidity,
factory_temperature, factory_humidity, and class",
    "parameters": {
        "type": "object",
        "properties": {
            "machine_temperature": {
                "description": "The temperature of the machine",
                "type": ["number", "null"]
            },
            "machine_humidity": {
                "description": "The humidity of the machine",
                "type": ["number", "null"]
            },
            "factory_temperature": {
                "description": "The temperature of the factory",
                "type": ["number", "null"]
            },
            "factory_humidity": {
                "description": "The humidity of the factory",
                "type": ["number", "null"]
            },
            "class": {
```

```
            "description": "Must be 0 or 1 or None; Good product: 0 / Defective product: 1 /
Unknown: None",
            "type": ["number", "null"],
            "enum": [0, 1, null]
          }
        },
        "required": [
          "machine_temperature",
          "machine_humidity",
          "factory_temperature",
          "factory_humidity",
          "class"
        ]
    }
}
```

## Supplementary Note 5. Multilingual Input Support and Validation

To evaluate the multilingual capabilities of the IM-Chat framework, we conducted a preliminary assessment across four languages commonly encountered in industrial settings, Korean, English, Thai, and Vietnamese. These languages were selected in consideration of the linguistic diversity observed in injection molding factories in South Korea, where many foreign workers are employed. Representative queries corresponding to each of the key tool categories were selected, and the system's performance was validated to ensure robust handling of multilingual inputs across all supported modules.

This evaluation focused on representative queries for each tool type, as the primary objective was to verify the correct operation of the multilingual pipeline rather than to conduct exhaustive benchmarking. Given that IM-Chat translates all inputs into English prior to reasoning and tool invocation, the system's behavior for each language is expected to be largely consistent once the translation is verified to be accurate. Accordingly, validation was conducted for four core functionalities, troubleshooting table retrieval, manufacturing manual retrieval, diffusion model invocation, and internet search, across four languages (Korean, English, Thai, and Vietnamese) to ensure robust multilingual support. The results, as shown in the tables below, indicate that IM-Chat produced consistent outcomes across different languages, demonstrating the robustness of its multilingual processing pipeline.

### 5.1 Task Requiring Knowledge from Troubleshooting Table

| English | How should Injection speed be adjusted to reduce burr defects? | Specifically, you should prioritize reducing Injection Speed 1, followed by Injection Speed 2, and then Injection Speed 3. |
| --- | --- | --- |
| Korean | 버 결함을 줄이기 위해 사출속도는 어떻게 조정해야 하나요? | 버 결함을 줄이기 위해 주입 속도(injection speed)를 조정하는 방법에 대한 조언은 다음과 같습니다:<br><br>주입 속도 3: 감소 (우선순위: 5)<br>주입 속도 2: 감소 (우선순위: 6)<br>주입 속도 1: 감소 (우선순위: 7) |
| Thai | ควรปรับความเร็วการฉีดอย่างไรเพื่อลดข้อบกพร่องของเสี้ยน (burr)? | เพื่อปรับลดข้อบกพร่องของเสี้ยน (burr defects) ควรปรับลดความเร็วในการฉีด (injection speed) โดยเริ่มจากการลดความเร็วในการฉีดที่ระดับ 3 (injection speed 3) ตามด้วยระดับ 2 (injection speed 2) และระดับ 1 (injection speed 1) ค่ะ |
| Vietnamese | Tốc độ ép phun nên được điều chỉnh như thế nào để giảm lỗi bavia? | Để giảm thiểu khuyết tật flash, bạn nên điều chỉnh tốc độ phun (injection speed) như sau:<br><br>Tốc độ phun 3: Giảm (Ưu tiên: 5)<br>Tốc độ phun 2: Giảm (Ưu tiên: 6)<br>Tốc độ phun 1: Giảm (Ưu tiên: 7) |

## 5.2 Task Requiring Knowledge from Manufacturing Manual

| English | What is the recommended Mold temperature range when the material is Acrylonitrile Butadiene Styrene (ABS)? | The recommended mold temperature range for Acrylonitrile Butadiene Styrene (ABS) is 40~60℃. |
|---|---|---|
| Korean | 사출하려는 재료가 Acrylonitrile Butadiene Styrene (ABS)일 때, 권장되는 몰드 온도범위는 어떻게 되나요? | ABS 재료의 권장 금형 온도 범위는 40~60℃입니다. |
| Thai | เมื่อใช้วัสดุฉีดขึ้นรูปเป็น Acrylonitrile Butadiene Styrene (ABS) ช่วงอุณหภูมิของแม่พิมพ์ที่แนะนำคือเท่าใด? | ช่วงอุณหภูมิแม่พิมพ์ที่แนะนำสำหรับการฉีดขึ้นรูป (injection molding) ของ Acrylonitrile Butadiene Styrene (ABS) คือ 40~60℃. |
| Vietnamese | Khi sử dụng vật liệu ép phun là Acrylonitrile Butadiene Styrene (ABS), dải nhiệt độ khuôn được khuyến nghị là bao nhiêu? | Nhiệt độ khuôn khuyến nghị cho vật liệu Acrylonitrile Butadiene Styrene (ABS) là 40~60℃. |

## 5.3 Task Requiring Knowledge from Diffusion Model

<table>
<tr>
<td>English</td>
<td>When Machine temperature is 20.5°C, Machine humidity is 42.0%, Factory temperature is 24.0°C, and Factory humidity is 36.0%, can you calculate the acceptable product conditions using the Diffusion model?</td>
<td>Based on the reasoning steps, the injection molding process parameter sets generated using the diffusion model with the given environmental conditions are as follows:<br><br>Injection Speed 1: 20.0<br>Injection Speed 2: 20.0<br>Injection Speed 3: 30.0<br>Injection Pressure 1: 120.0<br>Injection Pressure 2: 140.0<br>Injection Pressure 3: 130.0<br>Injection Position 1: 44.0<br>Injection Position 2: 32.0<br>Injection Position 3: 28.0<br>Hold Time: 0.0<br>These parameters are calculated considering the machine temperature of 20.5°C, machine humidity of 42.0%, factory temperature of 24.0°C, and factory humidity of 36.0%.</td>
</tr>
<tr>
<td>Korean</td>
<td>머신 온도가 20.5°C, 머신 습도가 42.0%, 공장 온도가 24.0°C, 공장 습도가 36.0%일 때, 디퓨전 모델로 양품조건을 계산해줄 수 있나요?</td>
<td>주어진 환경 조건을 사용하여 확산 모델(diffusion model)로 생성된 사출 성형(injection molding) 공정 매개변수 세트는 다음과 같습니다:<br><br>사출 속도 1: 38.00<br>사출 속도 2: 20.00<br>사출 속도 3: 30.00<br>사출 압력 1: 120.00<br>사출 압력 2: 120.00<br>사출 압력 3: 130.00<br>사출 위치 1: 44.00<br>사출 위치 2: 32.01<br>사출 위치 3: 28.00<br>보유 시간: 0.90 초</td>
</tr>
</table>

<table>
<tr><td></td><td></td><td>이 정보가 사용자의 요청에 대한 답변입니다.</td></tr>
<tr><td>Thai</td><td>เมื่ออุณหภูมิเครื่องจักรอยู่ที่ 20.5°C<br>ความชื้นของเครื่องจักรอยู่ที่ 42.0%<br>อุณหภูมิในโรงงานอยู่ที่ 24.0°C<br>และความชื้นในโรงงานอยู่ที่ 36.0%<br>โปรดคำนวณเงื่อนไขการผลิตที่ยอมรับได้โดยใช้แบบจำลองการแพร่ (Diffusion model)</td><td>ชุดพารามิเตอร์สำหรับกระบวนการฉีดขึ้นรูปที่สร้างขึ้นโดยใช้แบบจำลองการแพร่ (diffusion model) กับตัวแปรสิ่งแวดล้อมที่ให้มา มีดังนี้:<br><br>ความเร็วการฉีด (Injection Speed) 1: 20.0<br>ความเร็วการฉีด (Injection Speed) 2: 20.0<br>ความเร็วการฉีด (Injection Speed) 3: 30.0<br>แรงดันการฉีด (Injection Pressure) 1: 120.0<br>แรงดันการฉีด (Injection Pressure) 2: 120.0<br>แรงดันการฉีด (Injection Pressure) 3: 150.0<br>ตำแหน่งการฉีด (Injection Position) 1: 44.0<br>ตำแหน่งการฉีด (Injection Position) 2: 32.0<br>ตำแหน่งการฉีด (Injection Position) 3: 28.0<br>เวลาการถือ (Hold Time): 0.0</td></tr>
<tr><td>Vietnamese</td><td>Nếu nhiệt độ máy là 20.5°C, độ ẩm máy là 42.0%, nhiệt độ nhà máy là 24.0°C và độ ẩm nhà máy là 36.0%, thì mô hình khuếch tán có thể tính toán điều kiện sản xuất tối ưu không?</td><td>Mô hình khuếch tán (diffusion model) đã được sử dụng để tạo ra các bộ thông số quy trình ép phun (injection molding process parameter sets) dựa trên các biến môi trường đã cung cấp: Nhiệt độ máy 20.5°C, Độ ẩm máy 42.0%, Nhiệt độ nhà máy 24.0°C, và Độ ẩm nhà máy 36.0%. Các bộ thông số quy trình ép phun được tạo ra như sau:<br><br>Tốc độ phun 1 (Injection Speed 1): 20.0<br>Tốc độ phun 2 (Injection Speed 2): 20.0<br>Tốc độ phun 3 (Injection Speed 3): 30.0<br>Áp suất phun 1 (Injection Pressure 1): 120.0<br>Áp suất phun 2 (Injection Pressure 2): 120.0<br>Áp suất phun 3 (Injection Pressure 3): 150.0<br>Vị trí phun 1 (Injection Position 1): 44.0<br>Vị trí phun 2 (Injection Position 2): 32.0<br>Vị trí phun 3 (Injection Position 3): 28.0<br>Thời gian giữ (Hold Time): 0.0</td></tr>
</table>

### 5.4 Task Requiring Knowledge from Internet Search

| English | Which are the leading injection molding machine manufacturers in Japan? | Japan is home to several leading manufacturers of injection molding machines, known for their high-quality engineering and advanced technology. Here are some of the top companies in this field:<br><br>Sumitomo (SHI) Demag: A prominent player in the injection molding industry, Sumitomo offers a wide range of machines known for their precision and efficiency.<br><br>Nissei Plastic Industrial Co., Ltd.: Nissei is renowned for its innovative solutions and a broad lineup of injection molding machines that cater to various industries.<br><br>Toshiba Machine Co., Ltd.: Now known as Shibaura Machine, this company provides advanced injection molding machines with a focus on energy efficiency and high performance.<br><br>Fanuc Corporation: Known for its automation and robotics, Fanuc also produces high-quality injection molding machines that integrate advanced control systems.<br><br>JSW (The Japan Steel Works, Ltd.): JSW is a major manufacturer offering a range of injection molding machines known for their durability and precision.<br><br>Toyo Machinery & Metal Co., Ltd.: Toyo is recognized for its all-electric injection molding machines, which are designed for energy efficiency and high precision. |
|---|---|---|
| | | These companies are leaders in the industry due to their commitment to innovation, quality, and customer satisfaction. |
| Korean | 일본의 주요 사출성형기 제조사는 어디인가요? | 일본은 첨단 기술과 고품질 생산으로 유명한 여러 주요 사출 성형기 제조업체의 본고장입니다. 다음은 주요 제조업체들입니다:<br><br>Sumitomo (SHI) Demag: 이 분야의 선두 기업으로, Sumitomo 는 다양한 사출 성형기를 제공하며 혁신적인 기술과 에너지 효율적인 솔루션으로 잘 알려져 있습니다.<br><br>Nissei Plastic Industrial Co., Ltd.: Nissei 는 고성능 사출 성형기로 유명하며 국내외 시장에서 강력한 입지를 가지고 있습니다.<br><br>Toshiba Machine Co., Ltd.: 현재는 Shibaura Machine 으로 알려진 이 회사는 다양한 사출 성형기를 제공하며 정밀성과 신뢰성으로 인정받고 있습니다.<br><br>Fanuc Corporation: 자동화 및 로봇 공학으로 유명한 Fanuc 는 고효율적이고 첨단 제어 시스템을 통합한 사출 성형기도 생산합니다.<br><br>JSW (The Japan Steel Works, Ltd.): JSW 는 내구성과 첨단 기술로 유명한 사출 성형기를 제공하는 주요 업체입니다.<br><br>Toyo Machinery & Metal Co., Ltd.: Toyo 는 정밀성과 에너지 효율성으로 인기가 있는 전기식 사출 성형기로 잘 알려져 있습니다. |
| | | 이들 회사는 혁신과 품질에 중점을 두고 다양한 산업적 요구를 충족시키는 사출 성형기를 제공함으로써 사출 성형 산업에 크게 기여하고 있습니다. |
| Thai | ผู้ผลิตเครื่องฉีดพลาสติกหลักของญี่ปุ่นมีใครบ้าง? | ญี่ปุ่นเป็นที่ตั้งของผู้ผลิตเครื่องฉีดพลาสติกชั้นนำหลายราย นี่คือรายชื่อบริษัทชั้นนำในสาขานี้:<br><br>Sumitomo (SHI) Demag: เป็นที่รู้จักในด้านเทคโนโลยีขั้นสูงและเครื่องจักรคุณภาพสูง Sumitomo มีเครื่องฉีดพลาสติกหลากหลายรุ่น<br><br>Nissei Plastic Industrial Co., Ltd.: Nissei เป็นบริษัทที่มีชื่อเสียงซึ่งให้บริการเครื่องฉีดพลาสติกหลากหลายประเภทที่มีความแม่นยำและประสิทธิภาพสูง<br><br>Toshiba Machine Co., Ltd.: ปัจจุบันรู้จักกันในชื่อ Shibaura Machine พวกเขามีเครื่องฉีดพลาสติกหลากหลายรุ่นที่เหมาะสำหรับการใช้งานต่างๆ<br><br>Fanuc Corporation: Fanuc มีชื่อเสียงในด้านเทคโนโลยีอัตโนมัติและผลิตเครื่องฉีดพลาสติกไฟฟ้าที่มีประสิทธิภาพและความน่าเชื่อถือสูง<br><br>JSW (The Japan Steel Works, Ltd.): JSW เป็นผู้เล่นหลักในอุตสาหกรรมนี้ โดยมีเครื่องฉีดพลาสติกที่มีความทนทานและประสิทธิภาพสูง<br><br>Toyo Machinery & Metal Co., Ltd.: Toyo เชี่ยวชาญในเครื่องฉีดพลาสติกไฟฟ้าทั้งหมด ซึ่งมีความประหยัดพลังงานและความแม่นยำสูง |
| | | Ube Machinery Corporation, Ltd.: Ube ให้บริการเครื่องฉีดพลาสติกขนาดใหญ่และเป็นที่รู้จักในด้านโซลูชันนวัตกรรมในอุตสาหกรรม<br><br>บริษัทเหล่านี้ได้รับการยอมรับในด้านความก้าวหน้าทางเทคโนโลยีและการมีส่วนร่วมในอุตสาหกรรมการฉีดพลาสติกทั่วโลก |
| Vietnamese | Những nhà sản xuất máy ép phun chính của Nhật Bản là ai? | Nhật Bản là quê hương của nhiều nhà sản xuất máy ép phun nổi tiếng, được biết đến với kỹ thuật chất lượng cao và sự đổi mới. Dưới đây là một số nhà sản xuất chính:<br><br>Sumitomo (SHI) Demag: Là một công ty hàng đầu trong lĩnh vực này, Sumitomo cung cấp một loạt các máy ép phun, bao gồm các mẫu máy điện hoàn toàn, hybrid và thủy lực. Họ nổi tiếng với độ chính xác và hiệu quả năng lượng.<br><br>Nissei Plastic Industrial Co., Ltd.: Nissei nổi tiếng với công nghệ tiên tiến và dòng sản phẩm đa dạng, bao gồm các máy ép phun điện hoàn toàn, hybrid và thủy lực. Họ tập trung vào công nghệ bền vững và tiết kiệm năng lượng.<br><br>Toshiba Machine Co., Ltd.: Hiện nay được biết đến với tên Shibaura Machine, công ty này cung cấp nhiều loại máy ép phun được sử dụng trong các ngành công nghiệp khác nhau. Họ nhấn mạnh vào độ chính xác và độ tin cậy của máy móc.<br><br>Fanuc Corporation: Được biết đến với các giải pháp robot và tự động hóa, Fanuc cũng sản xuất các máy ép phun điện hoàn toàn. Máy của họ được công nhận về tốc độ cao, độ chính xác và hiệu quả năng lượng. |
|  |  | JSW (The Japan Steel Works, Ltd.): JSW cung cấp một loạt các máy ép phun, bao gồm các mẫu máy quy mô lớn cho các ứng dụng công nghiệp. Họ nổi tiếng với thiết kế chắc chắn và công nghệ tiên tiến.<br><br>Ube Machinery Corporation, Ltd.: Ube chuyên về các máy ép phun quy mô lớn và nổi tiếng với các giải pháp sáng tạo trong các ngành công nghiệp ô tô và công nghiệp.<br><br>Những công ty này là những người dẫn đầu trong ngành công nghiệp ép phun, cung cấp các máy móc được sử dụng trên toàn cầu trong nhiều lĩnh vực khác nhau, bao gồm ô tô, điện tử và hàng tiêu dùng. |

## 5.5 Task Requiring Knowledge from Troubleshooting Table and Diffusion Model

| English | When the Machine temperature is 22.0°C, Machine humidity is 43.5%, Factory temperature is 26.0°C, and Factory humidity is 34.0%, generate process parameters using the Diffusion model. Then, assuming a short shot defect occurs under these conditions, tell me how the Injection speed should be set based on the reference table. | Based on the given conditions, the process parameters generated using the Diffusion model are:<br><br>- Injection Speed 1: 20.0<br>- Injection Speed 2: 20.0<br>- Injection Speed 3: 30.0<br>- Injection Pressure 1: 120.0<br>- Injection Pressure 2: 120.0<br>- Injection Pressure 3: 130.0<br>- Injection Position 1: 44.0<br>- Injection Position 2: 32.0<br>- Injection Position 3: 32.0<br>- Hold Time: 0.0<br><br>In the event of a short shot defect, the recommended adjustments for the Injection speed are:<br><br>1. Increase Injection Speed 3 (Priority: 5)<br>2. Increase Injection Speed 2 (Priority: 6)<br>3. Increase Injection Speed 1 (Priority: 7)<br><br>These adjustments are based on senior worker-documented process parameter adjustments for addressing short shot defects. |
|---|---|---|
| Korean | 기계 온도가 22.0°C, 기계 습도가 43.5%, 공장 온도가 26.0°C, 공장 습도가 34.0°C 일 때, Diffusion 모델을 사용하여 공정 파라미터를 생성하시오. 그런 다음, 이러한 조건에서 short shot 결함이 발생했다고 가정하고, 참조 표를 기반으로 사출 속도(Injection speed) 를 어떻게 설정해야 하는지 알려주시오. | 주어진 환경 조건에서 확산 모델(diffusion model)을 사용하여 생성된 공정 매개변수는 다음과 같습니다:<br>주입 속도 1 (Injection Speed 1): 45.0<br>주입 속도 2 (Injection Speed 2): 45.0<br>주입 속도 3 (Injection Speed 3): 30.0<br>주입 압력 1 (Injection Pressure 1): 140.0<br>주입 압력 2 (Injection Pressure 2): 140.0 |
| | | 주입 압력 3 (Injection Pressure 3): 150.0<br>주입 위치 1 (Injection Position 1): 48.0<br>주입 위치 2 (Injection Position 2): 32.14<br>주입 위치 3 (Injection Position 3): 28.0<br>유지 시간 (Hold Time): 2.4 초<br>숏샷 결함(short shot defect)을 해결하기 위해, 선임 작업자가 문서화한 공정 매개변수 조정은 다음과 같이 주입 속도를 증가시키는 것을 제안합니다:<br><br>주입 속도 3 (Injection Speed 3): 증가 (우선순위: 5)<br>주입 속도 2 (Injection Speed 2): 증가 (우선순위: 6)<br>주입 속도 1 (Injection Speed 1): 증가 (우선순위: 7) |
| Thai | เมื่ออุณหภูมิเครื่องจักรอยู่ที่ 22.0°C ความชื้นของเครื่องจักรอยู่ที่ 43.5% อุณหภูโรงงานอยู่ที่ 26.0°C และความชื้นในโรงงานอยู่ที่ 34.0% ให้สร้างพารามิเตอร์กระบวนการโดยใช้แบบจำลอง Diffusion จากนั้นสมมติว่ามีข้อบกพร่องแบบ short shot เกิดขึ้นภายใต้เงื่อนไขเหล่านี้ โปรดบอกวิธีตั้งค่า Injection speed ตามตารางอ้างอิง | จากข้อมูลที่ได้รับมา<br>การแก้ไขปัญหาการเกิดข้อบกพร่องแบบ short shot สามารถทำได้โดยการปรับพารามิเตอร์กระบวนการตามลำดับความสำคัญดังนี้:<br><br>เพิ่มแรงดันการฉีด (Injection pressure) 3<br>เพิ่มแรงดันการฉีด (Injection pressure) 2<br>เพิ่มแรงดันการฉีด (Injection pressure) 1<br>เพิ่มอุณหภูมิของกระบอก (Barrel temperature)<br>เพิ่มความเร็วการฉีด (Injection speed) 3<br>เพิ่มความเร็วการฉีด (Injection speed) 2<br>เพิ่มความเร็วการฉีด (Injection speed) 1<br>เพิ่มแรงดันย้อนกลับ (Back pressure)<br>เพิ่มเวลาในการถือ (Hold time)<br>เพิ่มแรงดันในการถือ (Hold pressure) |
| | | พารามิเตอร์กระบวนการที่สร้างขึ้นโดยใช้โมเดลการแพร่ (Diffusion model)<br>ภายใต้สภาพแวดล้อมที่กำหนดมีดังนี้:<br><br>ความเร็วการฉีด (Injection Speed) 1: 20.0<br>ความเร็วการฉีด (Injection Speed) 2: 20.0<br>ความเร็วการฉีด (Injection Speed) 3: 30.0<br>แรงดันการฉีด (Injection Pressure) 1: 120.0<br>แรงดันการฉีด (Injection Pressure) 2: 120.0<br>แรงดันการฉีด (Injection Pressure) 3: 150.0<br>ตำแหน่งการฉีด (Injection Position) 1: 44.0<br>ตำแหน่งการฉีด (Injection Position) 2: 32.0<br>ตำแหน่งการฉีด (Injection Position) 3: 28.0<br>เวลาในการถือ (Hold Time): 0.0<br>ขอแนะนำให้ปรับพารามิเตอร์ตามลำดับความสำคัญที่กล่าวมาเพื่อแก้ไขปัญหา short shot ค่ะ |
| Vietnamese | Khi nhiệt độ của máy là 22.0°C, độ ẩm của máy là 43.5%, nhiệt độ của nhà máy là 26.0°C và độ ẩm của nhà máy là 34.0%, hãy tạo các tham số quy trình bằng mô hình Diffusion. Sau đó, giả sử có lỗi short shot xảy ra trong các điều kiện này, hãy cho tôi biết cách thiết lập tốc độ phun (Injection speed) dựa trên bảng tham chiếu. | Dựa trên các bước lập luận, thông tin cần thiết để trả lời yêu cầu của người dùng là:<br><br>Các thông số quy trình ép phun được tạo ra bằng cách sử dụng mô hình khuếch tán (diffusion model) với các biến môi trường đã cho như sau:<br><br>Tốc độ phun (Injection Speed) 1: 20.0<br>Tốc độ phun (Injection Speed) 2: 20.0<br>Tốc độ phun (Injection Speed) 3: 30.0<br>Áp suất phun (Injection Pressure) 1: 140.0<br>Áp suất phun (Injection Pressure) 2: 120.0<br>Áp suất phun (Injection Pressure) 3: 130.0<br>Vị trí phun (Injection Position) 1: 44.0<br>Vị trí phun (Injection Position) 2: 32.0<br>Vị trí phun (Injection Position) 3: 28.0<br>Thời gian giữ (Hold Time): 0.0 |

|  |  | Các điều chỉnh thông số quy trình do công nhân có kinh nghiệm ghi lại cho lỗi thiếu shot (short shot error) được sắp xếp theo thứ tự ưu tiên điều chỉnh như sau:<br><br>Áp suất phun 3: Tăng<br>Áp suất phun 2: Tăng<br>Áp suất phun 1: Tăng<br>Nhiệt độ thùng: Tăng<br>Tốc độ phun 3: Tăng<br>Tốc độ phun 2: Tăng<br>Tốc độ phun 1: Tăng<br>Áp suất ngược (Back pressure): Tăng<br>Thời gian giữ: Tăng<br>Áp suất giữ (Hold pressure): Tăng<br>Hy vọng thông tin này sẽ giúp bạn điều chỉnh tốc độ phun phù hợp với điều kiện nhiệt độ và độ ẩm hiện tại. |
|---|---|---|

## 5.6 Task Requiring Knowledge from Manufacturing Manual and Diffusion Model

| English | When Machine temperature is 22.0°C, Machine humidity is 43.5%, Factory temperature is 26.0°C, and Factory humidity is 34.0%, does the Injection pressure generated as a good condition in the Diffusion model fall within the allowed range for ABS according to the manual? | Based on the given environmental variables, the generated injection pressures in the diffusion model are 120.0, 120.0, and 133.26 ㎏ f/㎠. According to the manual, the allowed range for injection pressure for ABS is 750~1500 ㎏ f/㎠. Therefore, the generated injection pressures fall well within the allowed range for ABS. |
|---|---|---|
| Korean | 기계 온도가 22.0°C, 기계 습도가 43.5%, 공장 온도가 26.0°C, 공장 습도가 34.0°C 일 때, Diffusion 모델에서 양호한 조건으로 생성된 사출 압력(Injection pressure) 값이 매뉴얼에 명시된 ABS 의 허용 범위 내에 포함되는가? | 확산 모델(Diffusion model)이 생성한 주입 압력(Injection Pressure) 값은 120.0, 120.0, 130.0 입니다. ABS 의 허용 주입 압력 범위는 매뉴얼에 따르면 750~1500 ㎏ f/㎠입니다. |
| Thai | เมื่ออุณหภูมิเครื่องจักรอยู่ที่ 22.0°C<br>ความชื้นของเครื่องจักรอยู่ที่ 43.5%<br>อุณหภูมิในโรงงานอยู่ที่ 26.0°C<br>และความชื้นในโรงงานอยู่ที่ 34.0% ความดันการฉีด (Injection pressure)<br>ที่สร้างขึ้นภายใต้สภาวะดีโดยแบบจำลอง Diffusion นั้นอยู่ในช่วงที่อนุญาตสำหรับวัสดุ ABS<br>ตามคู่มือหรือไม่? | จากขั้นตอนการให้เหตุผล<br>ข้อมูลที่เกี่ยวข้องที่จำเป็นในการตอบคำถามคือ:<br><br>ชุดพารามิเตอร์กระบวนการฉีดขึ้นรูปที่สร้างขึ้นโดยใช้โมเดลการแพร่ (diffusion model)<br>ภายใต้สภาพแวดล้อมที่กำหนดมีแรงดันฉีด (Injection Pressure) ดังนี้:<br><br>แรงดันฉีด 1: 120.0<br>แรงดันฉีด 2: 120.0<br>แรงดันฉีด 3: 130.0<br>ช่วงแรงดันฉีดที่อนุญาตสำหรับวัสดุ ABS<br>ตามคู่มือเครื่องฉีดขึ้นรูปคือ 750~1500 ㎏ f/㎠<br><br>ดังนั้น<br>แรงดันฉีดที่สร้างขึ้นภายใต้สภาพแวดล้อมที่กำหนดอยู่ในช่วงที่อนุญาตสำหรับวัสดุ ABS<br>ตามคู่มือเครื่องฉีดขึ้นรูป |

| Vietnamese | Khi nhiệt độ của máy là 22.0°C, độ ẩm của máy là 43.5%, nhiệt độ của nhà máy là 26.0°C và độ ẩm của nhà máy là 34.0%, áp suất phun (Injection pressure) được tạo ra trong điều kiện tốt bởi mô hình Diffusion có nằm trong phạm vi cho phép đối với vật liệu ABS theo hướng dẫn sử dụng không? | Các thông số quy trình ép phun (injection molding process parameters) được tạo ra cho mô hình Diffusion, dựa trên các điều kiện môi trường đã cho, bao gồm áp suất phun (injection pressures) là 120.0, 120.0 và 150.0 ㎏ f/㎠. Theo hướng dẫn, phạm vi áp suất phun cho phép đối với vật liệu ABS là 750~1500 ㎏ f/㎠. Do đó, các áp suất phun được tạo ra bởi mô hình Diffusion nằm trong phạm vi cho phép đối với vật liệu ABS. |
|---|---|---|

## 5.7 Task Requiring Knowledge from Troubleshooting Table and Internet Search

| English | Use the predicted temperature and humidity at 2 PM tomorrow to set Factory Temperature and Factory Humidity, with Machine Temperature 1°C lower than Factory Temperature and Machine Humidity 5% higher than Factory Humidity. Then, input these into the Diffusion model to calculate the good conditions. | The predicted temperature at 2 PM tomorrow is 72°F, and the humidity is 72%. The Factory Temperature is set at 22.22°C (converted from 72°F), and the Factory Humidity is 72%. The Machine Temperature is calculated to be 21.22°C, which is 1°C lower than the Factory Temperature. The Machine Humidity is 75.6%, which is 5% higher than the Factory Humidity. These values are input into the diffusion model, which calculates the following optimal conditions for the injection molding process:<br><br>- Injection Speed 1: 20.0<br>- Injection Speed 2: 20.0<br>- Injection Speed 3: 30.0<br>- Injection Pressure 1: 120.0<br>- Injection Pressure 2: 120.0<br>- Injection Pressure 3: 130.0<br>- Injection Position 1: 44.21<br>- Injection Position 2: 32.0<br>- Injection Position 3: 28.0<br>- Hold Time: 2.4 seconds |
|---|---|---|
| Korean | 내일 오후 2 시의 기온과 습도를 Factory Temperature 와 Factory Humidity 로 설정하고,<br>Machine Temperature 는 Factory Temperature 보다 1°C 낮고,<br>Machine Humidity 는 Factory Humidity 보다 5% 높은 조건으로<br>Diffusion model 을 사용하여 양품 조건을 계산해주세요. | 주어진 환경 변수에 기반하여 확산 모델을 사용하여 생성된 사출 성형 공정 매개변수 세트는 다음과 같습니다:<br><br>사출 속도 (Injection Speed):<br><br>사출 속도 1: 45.0<br>사출 속도 2: 45.0<br>사출 속도 3: 30.0<br>사출 압력 (Injection Pressure):<br><br>사출 압력 1: 138.18 |
| | | 사출 압력 2: 140.0<br>사출 압력 3: 150.0<br>사출 위치 (Injection Position):<br><br>사출 위치 1: 48.0<br>사출 위치 2: 33.74<br>사출 위치 3: 28.0<br>보유 시간 (Hold Time): 2.40 초<br><br>이 정보는 주어진 조건 하에서 양호한 제품 조건을 계산하는 데 사용될 수 있습니다. |
| Thai | ตั้งค่าอุณหภูมิและความชื้นของโรงงาน (Factory Temperature และ Factory Humidity)<br>ให้เป็นค่าที่พยากรณ์ไว้ในวันพรุ่งนี้เวลา 14:00 น.<br>กำหนดให้อุณหภูมิของเครื่องจักร (Machine Temperature) ต่ำกว่าอุณหภูมิของโรงงาน 1°C<br>และความชื้นของเครื่องจักร (Machine Humidity)<br>สูงกว่าความชื้นของโรงงาน 5%<br>จากนั้นให้คำนวณเงื่อนไขสำหรับผลิตภัณฑ์ที่ดี (Good Condition) โดยใช้โมเดล Diffusion | จากขั้นตอนการให้เหตุผลที่ได้กล่าวถึง<br>การใช้โมเดลการแพร่ (diffusion model)<br>เพื่อสร้างชุดพารามิเตอร์สำหรับกระบวนการฉีดขึ้นรูป (injection molding)<br>โดยใช้ตัวแปรสิ่งแวดล้อมที่กำหนดไว้ ได้แก่<br>อุณหภูมิเครื่องจักร 23°C ความชื้นเครื่องจักร 56%<br>อุณหภูมิในโรงงาน 24°C และความชื้นในโรงงาน 51%<br>ชุดพารามิเตอร์ที่ได้มีดังนี้: ความเร็วการฉีด 1: 45.0<br>ความเร็วการฉีด 2: 20.0 ความเร็วการฉีด 3: 21.55<br>แรงดันการฉีด 1: 120.0 แรงดันการฉีด 2: 140.0<br>แรงดันการฉีด 3: 150.0 ตำแหน่งการฉีด 1: 48.0<br>ตำแหน่งการฉีด 2: 32.0 ตำแหน่งการฉีด 3: 32.0<br>เวลาการถือ: 2.4 วินาที<br>หวังว่าข้อมูลนี้จะเป็นประโยชน์ในการคำนวณสภาวะที่เหมาะสมสำหรับผลิตภัณฑ์ที่ดีค่ะ |
| Vietnamese | Thiết lập Nhiệt độ Nhà máy (Factory Temperature) và Độ ẩm Nhà máy (Factory Humidity) theo dự báo vào lúc 14:00 ngày mai.<br>Đặt Nhiệt độ Máy (Machine Temperature) thấp hơn Nhiệt độ Nhà máy 1°C và Độ ẩm Máy (Machine Humidity) cao hơn Độ ẩm Nhà máy 5%.<br>Sau đó, hãy sử dụng mô hình Diffusion để tính toán điều kiện đạt chuẩn (Good Condition). | Dựa trên các bước lập luận, thông tin cần thiết để trả lời yêu cầu là:<br><br>Dự báo thời tiết cho 14:00 ngày mai cho thấy nhiệt độ dao động từ 19°C đến 23°C và độ ẩm khoảng 48% tại San Jose.<br>Nhiệt độ của Máy được đặt thấp hơn 1°C so với Nhiệt độ của Nhà máy, tức là 18°C. |

| | | Độ ẩm của Máy được đặt cao hơn 5% so với Độ ẩm của Nhà máy, tức là 53%.<br>Sử dụng mô hình Khuếch tán (Diffusion model) với các thông số: Nhiệt độ Máy (18°C), Độ ẩm Máy (53%), Nhiệt độ Nhà máy (19°C), và Độ ẩm Nhà máy (48%) để tính toán các điều kiện tiêu chuẩn.<br>Mô hình khuếch tán đã tính toán các điều kiện tiêu chuẩn với các thông số đã cho như sau:<br><br>Tốc độ Tiêm 1: 20.0<br>Tốc độ Tiêm 2: 20.0<br>Tốc độ Tiêm 3: 30.0<br>Áp suất Tiêm 1: 120.0<br>Áp suất Tiêm 2: 120.0<br>Áp suất Tiêm 3: 130.0<br>Vị trí Tiêm 1: 44.0<br>Vị trí Tiêm 2: 32.0<br>Vị trí Tiêm 3: 28.36<br>Thời gian Giữ: 2.4 giây |
|---|---|---|

**Supplementary Note 6. Prompts for LLM-based Evaluation**

This prompt was used to configure the language model for conducting automated evaluations of IM-Chat's responses. Specifically, it enables self-assessment of the generated question–answer (Q&A) pairs by prompting the model to act as an expert evaluator. The model is instructed to assess each response along two key dimensions: (1) **Relevance**, which evaluates whether the answer addresses the user's intent, and (2) **Accuracy**, which checks the factual correctness of the information provided. For each criterion, the model is required to generate a brief textual justification, followed by a numerical score (0–10), where 10 denotes a highly relevant and accurate answer, and 0 indicates a completely inappropriate or incorrect response.

```
You are an expert evaluator reviewing the quality of an AI-generated answer.

Here is the user's original question:

{question}

Here is the answer provided by the AI:

{answer}

Evaluate the answer based on the following two criteria:

1. Relevance: Does the answer appropriately address the intent of the question?

2. Accuracy: Is the information factually correct based on general knowledge?
```

Provide a short evaluation for each criterion (1-2 sentences), followed by an overall score (0 to 10) where 10 means excellent and 0 means completely irrelevant or incorrect.

Format:

Relevance: ...

Accuracy: ...

Rating:0~10

## Supplementary Note 7. Dataset Construction and LLM Fine-Tuning

For implementing the fine-tuning of GPT-4o through the API, it was necessary to construct a training dataset that emulates conversations between the knowledge transfer system and manufacturing workers. To curate the query–response pairs for the troubleshooting table, manufacturing manual, and diffusion model, we followed the procedure described below.

The troubleshooting table contains, for each defect type, the corresponding process parameters that should be increased or decreased, along with the priority order in which these parameters should be adjusted. To collect query-response pairs reflecting this knowledge, we first defined a prompt format and then created ten distinct questions for each defect type based on the information in the table. This format was designed to elicit responses related to key process parameters and their adjustment strategies for defect resolution. Specifically, the questions were structured to identify which parameter is most critical for resolving a given defect, to request a step-by-step explanation of the corrective procedure, and to inquire how the adjustment of a specific parameter may contribute to mitigating certain defect types. The LLM (GPT-4o) was then prompted to generate answers to these questions, forming the query-response for fine-tuning. All generated responses were manually reviewed by human evaluators to ensure the accuracy of the training dataset.

For the manual retrieval, it was highly labor-intensive for humans to read all 227 pages of the manufacturing manual and manually construct questions-answer pairs. To mitigate this effort, the LLM was prompted to generate several question-answer pairs for each page. The generated pairs were then combined with those manually created by human experts, resulting in a dataset of 1,000 entries. As with the troubleshooting table dataset, all samples were manually reviewed to ensure their accuracy.

For the diffusion model dataset, the dataset used for diffusion model training in prior experiments was directly employed for fine-tuning in a conversational format. Specifically, the conversations included user queries about factory temperature and humidity, machine temperature and humidity, and whether the condition corresponded to a normal or defective product. In response, the system generated the corresponding process parameters, including injection speeds (1st, 2nd, and 3rd stages), injection pressures (1st, 2nd, and 3rd stages), injection positions (1st, 2nd, and 3rd stages), and hold time. A total of 2,797 conversational pairs were incorporated into the fine-tuning dataset.

The three datasets described above were merged to construct the final training dataset for fine-tuning. The fine-tuning of GPT-4o was then performed using the OpenAI API's fine-tuning functionality. The corresponding cross-entropy loss curve is presented in **Figure S2**.

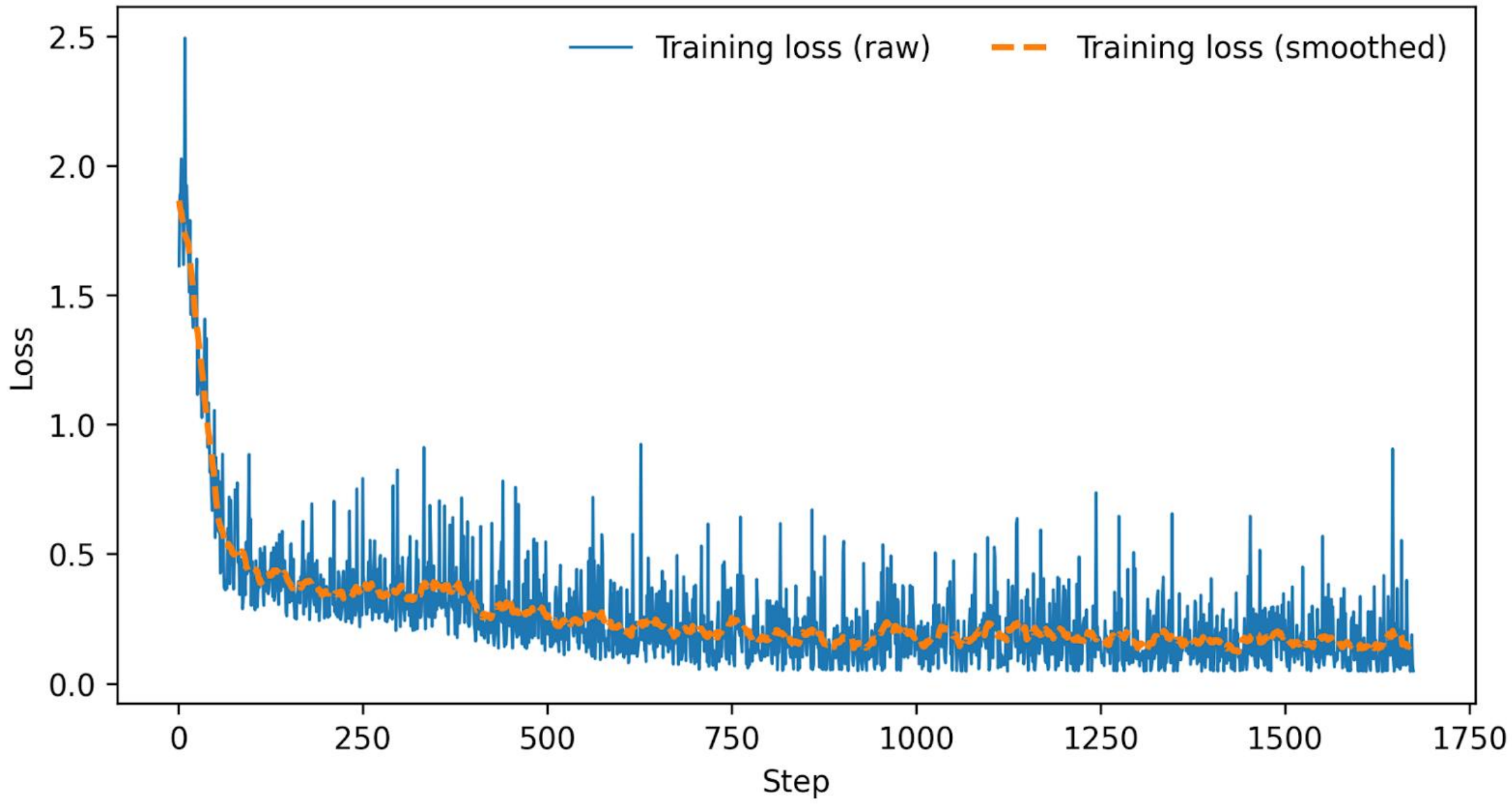

**Figure S2:** Training loss curve of the fine-tuned GPT-4o model